%% file: 0_main.tex
\documentclass[10pt,journal]{IEEEtran}
\usepackage[ruled,vlined]{algorithm2e}
\usepackage{graphicx}
\usepackage{booktabs}
\usepackage{stmaryrd}
\usepackage{tabularx}
\usepackage{subcaption}
\usepackage{amsmath,amssymb} 
\usepackage{multirow}
\usepackage{color}
\usepackage[switch]{lineno}

\def\ie{{\it i.e.}}
\def\eg{{\it e.g.}}
\def\et{{\it et al.~}}
\def\etc{{\it etc.}}

\def\I{{\mathbf I}}
\def\A{{\mathbf A}}
\def\W{{\mathbf W}}
\def\P{{\mathbf P}}

\def\X{{\mathbf X}}
\def\Y{{\mathbf Y}}
\def\Z{{\mathbf Z}}

\SetKwInput{KwInput}{Input}
\SetKwInput{KwOutput}{Output}

\begin{document}

\title{AGRNet: Adaptive Graph Representation Learning and Reasoning for Face Parsing} 

\author{
    Gusi~Te,~\IEEEmembership{Student Member,~IEEE,}
    Wei~Hu,~\IEEEmembership{Senior Member,~IEEE,}
    Yinglu~Liu,~\IEEEmembership{Member,~IEEE,}
	Hailin~Shi,~\IEEEmembership{Member,~IEEE,}
	and~Tao~Mei,~\IEEEmembership{Fellow,~IEEE}
\thanks{G. Te and W. Hu are with Wangxuan Institute of Computer Technology, Peking University, No. 128, Zhongguancun North Street, Beijing, China. E-mail: \{tegusi,forhuwei\}@pku.edu.cn.}
\thanks{Y. Liu, H. Shi and T. Mei are with JD AI Research, Beijing, China. E-mail: \{liuyinglu1,shihailin,tmei\}@jd.com. }
\thanks{Corresponding author: Wei Hu (forhuwei@pku.edu.cn). }
\thanks{This work was supported by National Natural Science Foundation of China (61972009). }
}

\maketitle

\begin{abstract}
Face parsing infers a pixel-wise label to each facial component, which has drawn much attention recently.
Previous methods have shown their success in face parsing, which however overlook the correlation among facial components.
As a matter of fact, the component-wise relationship is a critical clue in discriminating ambiguous pixels in facial area.
To address this issue, we propose adaptive graph representation learning and reasoning over facial components, aiming to learn representative vertices that describe each component, exploit the component-wise relationship and thereby produce accurate parsing results against ambiguity. 
In particular, we devise an adaptive and differentiable graph abstraction method to represent the components on a graph via pixel-to-vertex projection under the initial condition of a predicted parsing map, where pixel features within a certain facial region are aggregated onto a vertex. 
Further, we explicitly incorporate the image edge as a prior in the model, which helps to discriminate edge and non-edge pixels during the projection, thus leading to refined parsing results along the edges.
Then, our model learns and reasons over the relations among components by propagating information across vertices on the graph. 
Finally, the refined vertex features are projected back to pixel grids for the prediction of the final parsing map.
To train our model, we propose a discriminative loss to penalize small distances between vertices in the feature space, which leads to distinct vertices with strong semantics.  
Experimental results show the superior performance of the proposed model on multiple face parsing datasets, along with the validation on the human parsing task to demonstrate the generalizability of our model.   

\end{abstract}

\begin{figure}[t]
    \centering
    \includegraphics[width=0.5\textwidth]{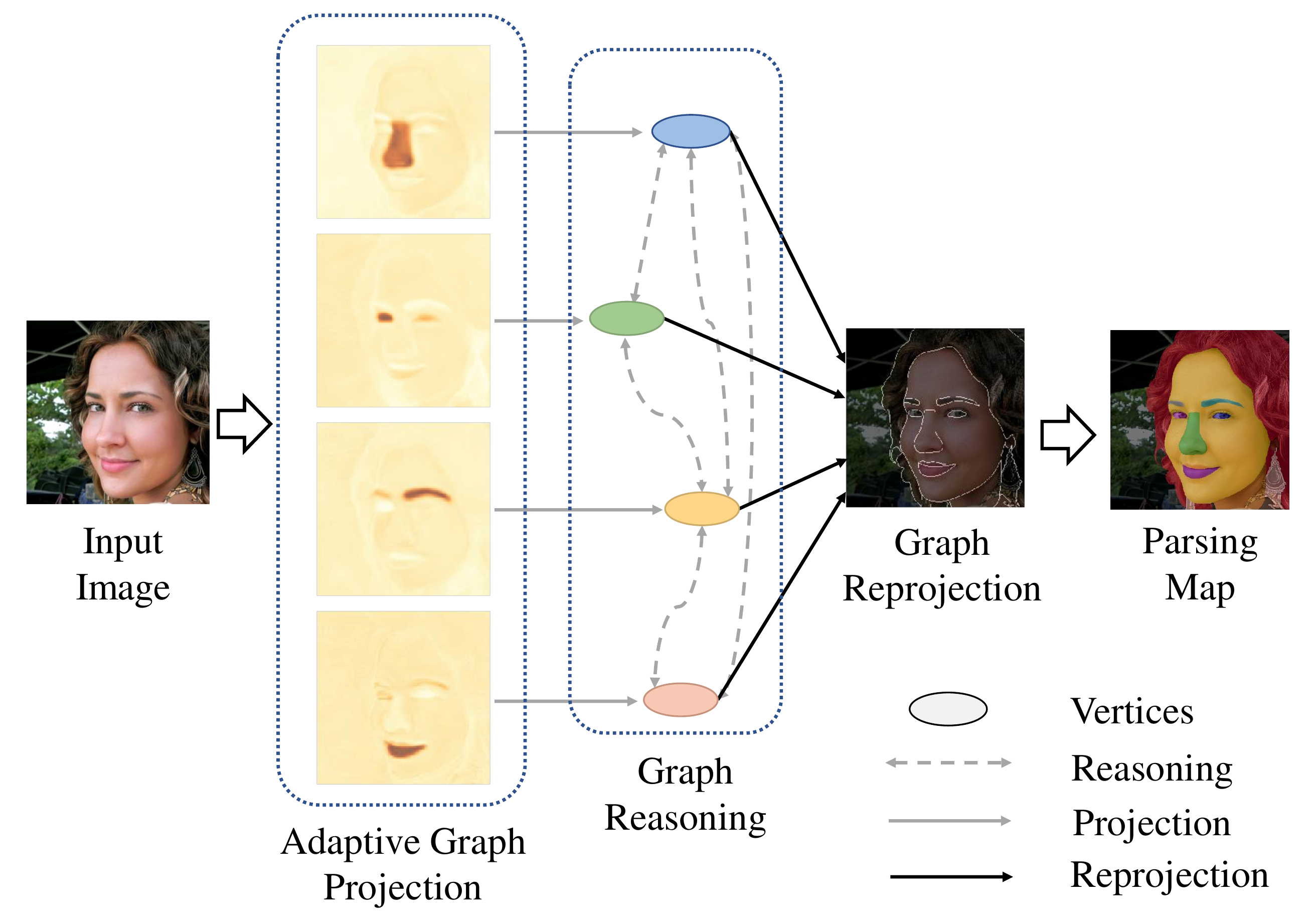}
    \caption{\textbf{Illustration of the proposed adaptive graph representation learning and reasoning for face parsing, which aims to capture the long range dependencies among facial components.} Given an input image, we represent each facial component by a few vertices via adaptive pixel-to-vertex projection, learn and reason over the graph connectivity to infer the relations among facial components, and reproject the refined vertex features to the pixel grids for face parsing.}
    \label{fig:teaser}
\end{figure}

\section{Introduction}
\label{sec:intro}
\input{1_intro.tex}

\section{Related Work}
\label{sec:related}

\input{2_related.tex}

\section{Adaptive Graph Representation Learning and Reasoning}
\label{sec:method}
\input{3_framework}

\section{Training Objectives and Analysis}
\label{sec:training}
\input{3_training}

\section{Experiments}
\label{sec:results}
\input{4_results}

\section{Conclusion}
\label{sec:conclude}
\input{5_conclusion}


\bibliographystyle{IEEEtran}
\bibliography{6_ref}
\newpage
\input{bio}

\end{document}

%% file: 1_intro.tex
Face parsing is a particular task in semantic segmentation, which assigns a pixel-wise label to each semantic component, such as facial skin, eyes, mouth and nose.
It enables more complex face analysis tasks, including face swapping \cite{KemelmacherShlizerman2016TransfiguringP,nirkin2018face}, face editing  \cite{shu2017neural, zhang2018synthesis} and face completion \cite{li2017generative,zhang2016joint}.

A plethora of methods have been proposed to fulfill this task with remarkable success, which can be classified into two categories: 
1) Face-based parsing, which takes the holistic face image as input for segmentation \cite{liu2015multi,wei2017learning,liu2020new,te2020edge, luo2020ehanet}. This however may neglect the scale discrepancy in different facial components, sometimes resulting in the lack of details. 
2) Region-based parsing, which addresses the above problem by first predicting the bounding box and then acquiring the parsing map of each facial component individually \cite{liu2017face,lin2019face, yin2020end, zhou2015interlinked,luo2012hierarchical}. 
However, there still exist limitations and challenges. 
Firstly, region-based parsing methods are based on the individual information within each component, while the {\it correlation} among components is not exploited yet to capture long range dependencies. 
In fact, facial components present themselves with abundant correlation between each other. 
For instance, eyes, mouth and eyebrows will generally become more curvy when people smile, as shown in Fig.~\ref{fig:teaser}; facial skin and other components will be dark when the lighting is weak, \etc.
Secondly, it remains a challenge to segment pixels around the boundary between components, since boundaries tend to be ambiguous in real scenarios.  

To this end, we explicitly model the component-wise correlations on a graph abstraction, with each vertex describing a facial component and each link capturing the correlation between a pair of components. 
To learn such a graph representation, we propose Adaptive Graph Representation learning and reasoning over facial components---AGRNet, which learns and reasons over non-local components to capture long range dependencies with the boundary information between components leveraged. 
In particular, to bridge the image pixels and graph vertices, AGRNet adaptively projects a collection of pixels with similar features to each vertex.
To obtain representative vertices, instead of projecting pixels within an image block to a vertex by fixed spatial pooling as in our previous work \cite{te2020edge}, we adaptively select pixels with high responses to a distinct facial component as a graph vertex. Since facial components are unknown in the beginning, we employ a predicted parsing map as the initial condition.  
Further, to achieve accurate segmentation along the edges between components, AGRNet incorporates edge attention in the pixel-to-vertex projection, which assigns larger weights to the features of edge pixels during the feature aggregation. 

Based on the projected vertices, AGRNet learns the correlations among facial components, \ie, the graph links between vertices, and reasons over the correlations by propagating information across all vertices on the graph via graph convolution \cite{kipf2016semi}. 
This enables characterizing long range correlations in the facial image for learning high-level semantic information. 
Finally, we project the learned graph representation back to the pixel grids, which is integrated with the original pixel-wise feature map for face parsing. 

To train the proposed model, we design a boundary-attention loss as well as a discriminative loss to reinforce edge pixels and vertex features respectively. 
The boundary-attention loss measures the error of the predicted parsing map only at edge pixels, which aims to obtain accurate segmentation along the boundary. 
Meanwhile, the discriminative loss is designed to penalize small distances among vertices in the feature space, thus leading to more representative and distinct vertices with strong semantics.

Compared with our previous work \cite{te2020edge}, the proposed AGRNet makes significant improvements in the following three aspects. 
1) AGRNet projects pixels to vertices in an {\it adaptive} fashion conditioned on a predicted parsing map, rather than simply applying spatial pooling in \cite{te2020edge}. Specifically, as illustrated in Fig. \ref{fig:pooling}, \cite{te2020edge} projects pixels within an image patch to a vertex via spatial pooling, which results in ambiguous semantics in vertices. In contrast, AGRNet projects a collection of pixels to a vertex adaptively so as to represent each semantic facial component, which leads to more semantic and compact vertices for the subsequent efficient reasoning among components. This improvement is substantial and critical for capturing the long-range dependencies between facial components. 
2) We design a discriminative loss, which encourages vertices to keep distant in the feature space. This is advantageous in learning distinct vertices for accurate segmentation of ambiguous pixels along the boundary. While the parsing map in our model provides hints in selecting semantic vertices, the similarity among different vertices is overlooked, thus the discriminative loss is critical in pushing pixels to its corresponding component.
3) We conduct more extensive experiments on large-scale datasets, including LaPa \cite{liu2020new} and CelebAMask-HQ \cite{CelebAMask-HQ}. Also, we generalize the proposed model to the human parsing task on the LIP dataset \cite{gong2017look}, and the results demonstrate the effectiveness of our model.

Our main contributions are summarized as follows.
\begin{enumerate}
    \item We propose a component-level adaptive graph representation learning and reasoning for face parsing---AGRNet, aiming to exploit the correlations among components for capturing long range dependencies.
    
    \item The graph representation is adaptively learned by projecting a collection of pixels with similar features to each vertex in a differentiable manner along with edge attention. 
    
    \item To train the proposed model effectively, we propose a discriminative loss, which conduces to enlarging the feature distance among distinct vertices.
    
    \item We conduct extensive experiments on the benchmarks of face parsing as well as human parsing to show the generalization. Results demonstrate that our model outperforms the state-of-the-art methods on almost every category for face parsing.
\end{enumerate}

The paper is organized as follows. 
We first review previous works in Section~\ref{sec:related}. 
Then we elaborate on the proposed model in Section~\ref{sec:method}. 
Next, we present the loss functions for training the proposed model and provide further analysis in Section~\ref{sec:training}.
Finally, experiments and conclusions are presented in Section~\ref{sec:results} and \ref{sec:conclude}, respectively.

%% file: 2_related.tex

\subsection{Face Parsing}


Previous methods can be classified into two categories: face-based parsing and region-based parsing.

\textbf{Face-based parsing} takes the whole image as input regardless of component properties \cite{liu2015multi,wei2017learning,liu2020new,te2020edge, luo2020ehanet}. 
Liu \et import features learned from a convolutional neural network into the Conditional Random Field (CRF) framework to model individual pixel labels and neighborhood dependencies \cite{liu2015multi}. Luo \et propose a hierarchical deep neural network to extract multi-scale facial features \cite{luo2012hierarchical}. 
Zhou \et adopt an adversarial learning approach to train the network and capture the high-order inconsistency \cite{zhou2013extensive}. 
Liu \et design a CNN-RNN hybrid model that benefits from both high-quality features of CNNs and non-local properties of RNNs \cite{liu2017face}. 
To refine the segmentation along critical edges, Liu \et introduce edge cues and propose a boundary-aware loss for face parsing, leading to improved performance \cite{liu2020new}.
However, this class of methods neglect the discrepancy in scales of various facial components, sometimes resulting in the lack of details in some components. 

\textbf{Region-based parsing} takes the scale discrepancy into account and predicts each component respectively, which is advantageous in capturing elaborate details \cite{liu2017face,lin2019face, yin2020end, zhou2015interlinked,luo2012hierarchical}. 
Zhou \et present an interlinked CNN that takes multi-scale images as input and allows bidirectional information passing \cite{zhou2015interlinked}. 
Lin \et propose a novel RoI Tanh-Warping operator that preserves both central and peripheral information. It contains two branches with the local-based branch for inner facial components and the global-based branch for outer facial ones. This method demonstrates high performance especially for hair segmentation \cite{lin2019face}. 
Yin \et introduce the Spatial Transformer Network and build a training connection between traditional interlinked CNNs, which makes the end-to-end joint training process possible \cite{yin2020end}.
Nevertheless, this class of methods often neglect the correlation among components to characterize long range dependencies.

\begin{figure*}[t]
    \centering
    \includegraphics[width=\textwidth]{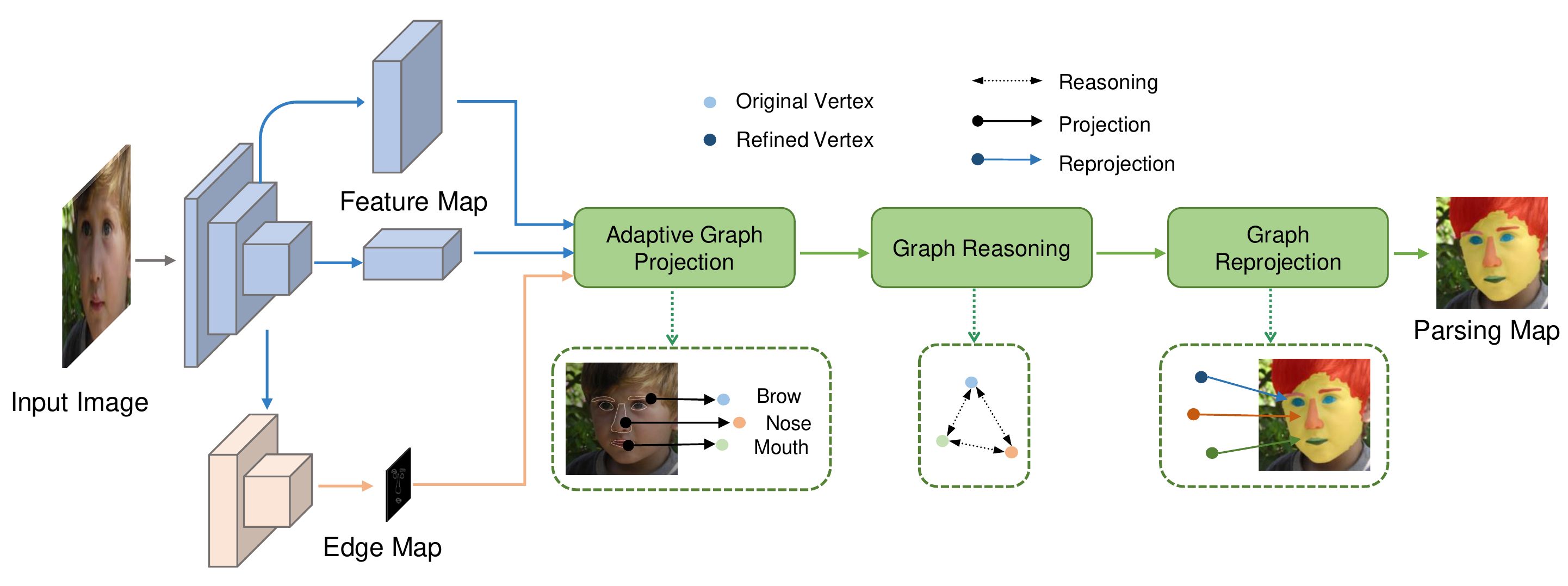}
    \caption{{The overview of the proposed face parsing framework.}}
    \label{fig:framework}
\end{figure*}

\subsection{Attention Mechanism}
Limited by the locality of the convolution operator, CNNs lack the ability to model the global contextual information in a single layer. Therefore, attention mechanism has been proposed to capture long-range information \cite{bahdanau2014neural}, which has been successfully applied to many applications such as sentence encoding \cite{vaswani2017attention} and image recognition \cite{wang2018non}. 

Chen \et propose a Double Attention Model that gathers information spatially and temporally to reduce the  complexity of traditional non-local modules \cite{chen20182}. 
Zhu \et present an asymmetric module to reduce the computation and distill features \cite{zhu2019asymmetric}. Fu \et devise a dual attention module that applies both spatial and channel attention in feature maps \cite{fu2019dual}. 
To research the underlying relationship among different regions, Chen \et project the original features into an interactive space and utilize the GCN \cite{kipf2016semi} to exploit the high-order relationship \cite{chen2019graph}. Li \et devise a robust attention module that incorporates the Expectation-Maximization algorithm \cite{li2019expectation}. Yin \et propose the disentangled non-local module, aiming at disentangling pairwise and unary relations in the classical non-local network \cite{yin2020disentangled}.

\subsection{Graph Representation for Images}
Images could be interpreted as a particular form of regular grid graphs, which thus enables studying images from the graph perspective. 
Chandra \et propose a Conditional Random Field based method on image segmentation \cite{chandra2017dense}. 
Recently, graph convolution networks \cite{kipf2016semi, defferrard2016convolutional,velivckovic2018graph} have been leveraged in image segmentation. 
Li \et introduce the graph convolution to semantic segmentation, which projects features onto vertices in the graph domain and applies graph convolution afterwards \cite{li2018beyond}. Furthermore, Lu \et propose Graph-FCN where semantic segmentation is cast as vertex classification by directly transforming an image into regular grids \cite{lu2019graph}. Pourian \et propose a graph-based method for semi-supervised segmentation \cite{pourian2015weakly}, where the image is divided into community graphs and labels are assigned to corresponding communities. Zhang \et utilize the graph convolution both in the coordinate space and the feature space \cite{zhang2019dual}. Li \et propose a spatial pyramid graph reasoning module based on an improved Laplacian matrix to learn a better distance metric for feature filtering \cite{li2020spatial}.

%% file: 3_framework.tex
In this section, we first introduce the overall framework in Section~\ref{subsec:overview}. 
Then, we elaborate on the proposed adaptive graph projection, graph reasoning, and graph reprojection in Section~\ref{subsec:projection}, Section~\ref{subsec:reasoning} and Section~\ref{subsec:reprojection}, respectively.  

\subsection{Overview}
\label{subsec:overview}

As illustrated in Fig.~\ref{fig:framework}, given an input face image $\mathbf{I} \in \mathbb{R}^{H \times W}$ of height $H$ and width $W$, we aim to predict a pixel-wise label to each facial semantic component. 
In particular, we aim to model the long-range dependencies among distant components on a graph, which is critical for the description of the facial structure. 
The overall framework of the proposed AGRNet consists of three procedures as follows. 

\begin{itemize}
    \item \textbf{Initial Feature and Edge Extraction.} We take the ResNet-101 \cite{he2016deep} as the backbone and extract low-level and high-level features for the subsequent graph representation. 
    The low-level features generally contain image details but often lack semantic information, while the high-level features provide rich semantics with global information at the cost of image details. 
    To fully exploit the global information in high-level features, we employ spatial pyramid pooling to learn multi-scale contextual information. 
    Further, we propose an edge perceiving module to acquire an edge map for the subsequent edge attention operation. 
    
    \item \textbf{Adaptive Graph Representation Learning and Reasoning.} We project a collection of pixels to similar features to graph vertices in an adaptive and differentiable manner, and reason over the vertices to learn the long-range relations among facial components. This consists of three operations: {\it adaptive graph projection}, {\it graph reasoning} and {\it graph reprojection}, which extracts component features as vertices in a differentiable fashion, reasons the relations between vertices with a graph convolution network, and projects the learned graph representation back to pixel grids, leading to a refined feature map with rich local and non-local semantics. 
    
    \item \textbf{Semantic Decoding.} Finally, we add the refined feature map to the original pixel-wise feature map and predict the parsing map via a convolution layer with kernel size of $1\times1$ .  
\end{itemize}

\begin{figure*}[t]
    \centering
    \includegraphics[width=\textwidth]{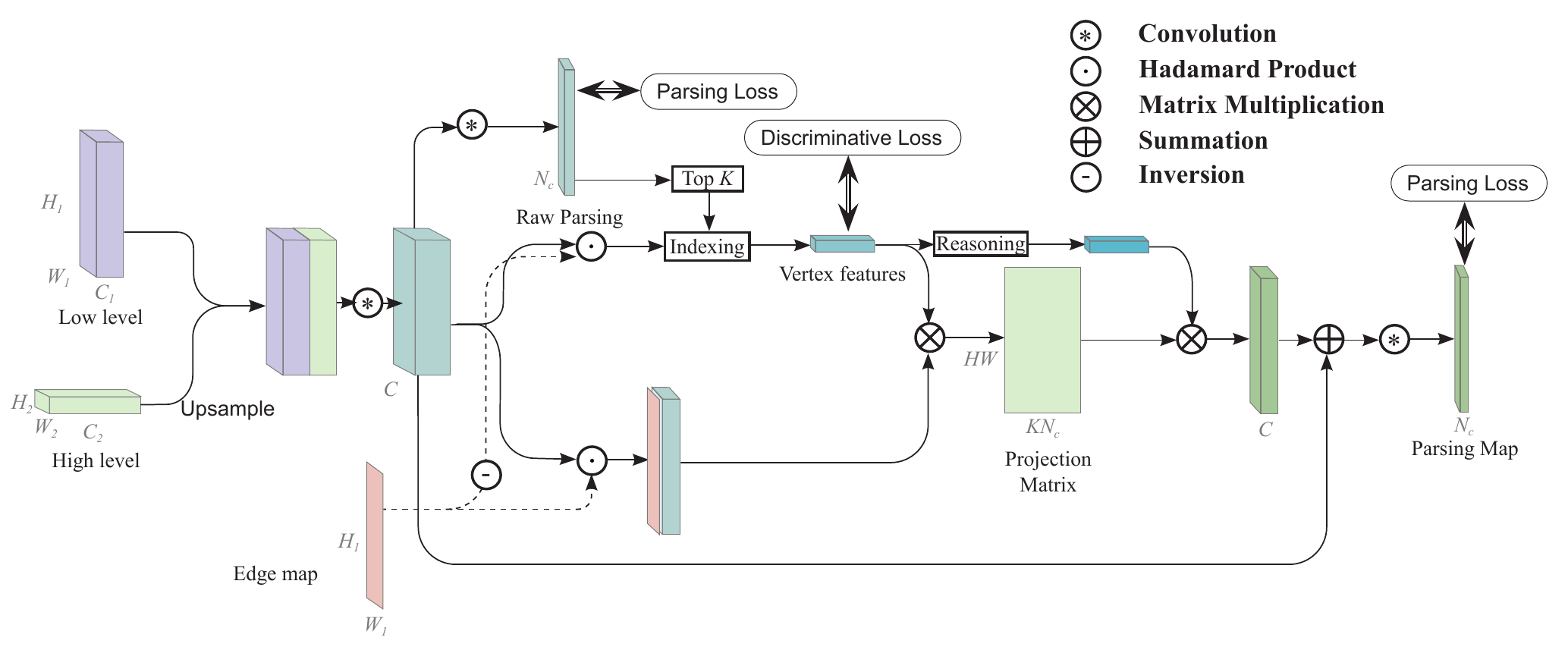}
    \caption{{Architecture of the proposed adaptive graph representation learning and reasoning for face parsing.}}
        \label{fig:module}
\end{figure*}

Next, we will elaborate on the three operations in {\it Adaptive Graph Representation Learning and Reasoning} in order as follows. 
The specific architecture is illustrated in Fig.~\ref{fig:module}.

\begin{figure}[t]
    \centering
    \includegraphics[width=0.44\textwidth]{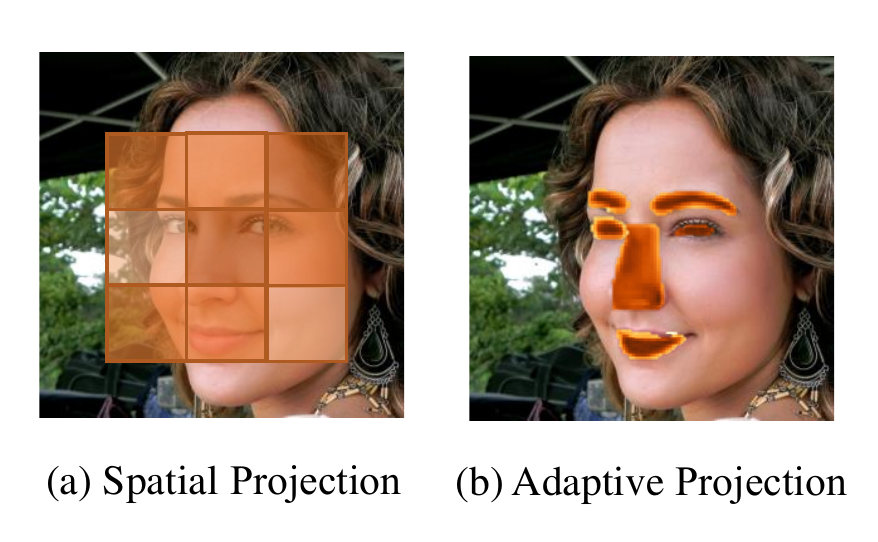}
    \vspace{-0.1in}
    \caption{\textbf{Comparison of the spatial graph projection in \cite{te2020edge} and the proposed adaptive graph projection.} (a) Spatial graph projection, where pixels within an image patch are projected to a vertex via spatial pooling. (b) Adaptive graph projection, where a collection of pixels are projected to a vertex adaptively so as to represent each semantic facial component.  }
    \label{fig:pooling}
\end{figure}

\subsection{Adaptive Graph Projection}
\label{subsec:projection}

We learn the graph representation of facial components in an {\it adaptive} and {\it differentiable} manner, which projects a collection of pixels that tend to reside in the same facial component to $K$ ($K \geq 1$) vertices in the graph. 
We choose $K$ vertices to represent a facial component, because 1) the feature from K vertices provides abundant visual description of the component, and 2) both the {\it intra-component} relations and {\it inter-component relations} will be exploited during the graph reasoning.    

Considering the facial components are unknown in the test, we first generate a preliminary raw parsing map that provides rough component information, which is supervised by the ground truth parsing map during the training. Based on the raw prediction, we select $K$ pixels with the highest confidence with respect to each component. Then the corresponding indices are utilized to sample representative vertices on the original feature map. These vertices will be fed into the subsequent graph reasoning module to extract the mutual relationship.

Specifically, the input to the proposed adaptive graph projection consists of a low-level feature map $\X_1 \in \mathbb{R}^{H_1W_1 \times C_1}$ and a high-level feature map $\X_2 \in \mathbb{R}^{H_2W_2 \times C_2}$ ($H_1W_1 > H_2W_2$) as well as an edge map $\mathbf E \in \mathbb{R}^{H_1W_1 \times 1}$ from the first procedure of Initial Feature and Edge Extraction, where $H_i$ and $W_i$ denote the height and width of the $i$th feature map respectively while $C_i$ is the number of feature channels, $i\in\{1,2\}$. 
Taking them as input, we aim to construct a set of vertices, each of which corresponds to a distinct facial component.


First, we upsample the feature map $\X_2$ to the scale of $\X_1$ with bilinear interpolation, and concatenate the upsampled version $\widetilde{\X}_2$ and $\X_1$ followed by a 1 $\times$ 1 convolution layer to reduce the channel dimension to $C$. Then we obtain the fused feature map $\X_{0} \in \mathbb{R}^{H_1W_1 \times C}$, with both detailed texture and abundant semantic information:
\begin{equation}
\X_{0} = \text{conv}([\X_1, \widetilde{\X}_2]),
\end{equation}
where $[\cdot,\cdot]$ denotes the concatenation operation, and $\text{conv}(\cdot)$ represents the 1 $\times$ 1 convolution.   

To separate pixels into edge pixels and non-edge pixels, we employ the learned edge map $\mathbf{E}$ to mask the original feature map $\X_{0}$:
\begin{equation}
\begin{aligned}
\X_{e} &= \X_0 \circ \mathbf{E}, \\
\X_{ne} &= \X_0 \circ (\mathbf{1} - \mathbf{E}), \\
\end{aligned}
\end{equation}
where $\circ$ denotes the Hadamard product and $\mathbf{1}$ is a matrix with all the elements as $1$. $\X_{e}$ and $\X_{ne}$ denote the masked feature map of edge pixels and non-edge pixels, respectively.  

In order to generate vertices that capture semantic features of facial components, we propose to first predict a preliminary parsing map $\Z_0 \in \mathbb{R}^{H_1W_1 \times N_c}$, where $N_c$ denotes the number of segmentation categories. 
The prediction of $\Z_0$ is produced by a Multi-layer Perceptron (MLP) under the supervision of the ground truth parsing map, \ie, $\Z_0=\text{conv}(\X_0)$. 
The predicted parsing map $\Z_0$ corresponds to the confidence in terms of each facial component. 

As component vertices should be compact and representative, we choose top $K$ pixels with highest confidence in $\Z_0$ as component vertices. 
Specifically, we extract the top $K$ pixels and their indices, along with the corresponding feature vectors from the masked feature map of non-edge pixels $\X_{ne}$.  
That is, only non-edge interior pixels are taken from $\X_{ne}$ as vertices:
\begin{equation}
\begin{aligned}
\X_{G} &= \X_{ne}\{\bigparallel_{i=1}^{N_c}\text{topk}(\Z_0[:,i])\} \\
&=  \X_{ne}\{\bigparallel_{i=1}^{N_c}\text{topk}(\text{conv}(\X_0)[:,i])\},
\end{aligned}
\end{equation}
where $\{\cdot\}$ denotes selection by the indices. 
The semantic vertices are then effectively extracted, each of which corresponds to a vertex in the graph. 

By the adaptive graph projection, we bridge the connection between pixels and each component via the selected vertices, leading to the features of the projected vertices on the graph $\X_G \in \mathbb{R}^{KN_c \times C}$.

The adaptive graph projection significantly improves the projection in our previous work \cite{te2020edge}. As in Fig.~\ref{fig:pooling}, we improve the traditional regular spatial pooling strategy by adaptively selecting representative vertices. Although facial components tend to locate in certain positions, regular spatial pooling fails to capture the representative information in each component and introduces irrelevant pixels. 
In contrast, our model characterizes each component with vertices adaptively, leveraging on a predicted parsing map that provides confident semantic information.

\subsection{Graph Reasoning}
\label{subsec:reasoning}

The connectivity between vertices represents the relation between each pair of facial components. 
Hence, we reason over the relations by propagating information across vertices to learn higher-level semantic information. 
Instead of constructing a pre-defined graph, the proposed AGRNet learns the graph connectivity dynamically without the supervision of a ground truth graph. 
We propose to leverage a single-layer Graph Convolution Network (GCN) \cite{kipf2016semi}---a first-order approximation of spectral graph convolution---to aggregate the neighborhood information and learn local vertex features. 

Specifically, we feed the input vertex features $\X_G$ into the GCN. 
The output feature map $\hat{\X}_G \in \mathbb{R}^{KN_c \times C}$ is
\begin{equation}
    \hat{\X}_G = \text{ReLU}\left[(\I - \A)\X_G\W_G\right],
\end{equation}
where $\A$ denotes the adjacency matrix that encodes the graph connectivity to learn, $\W_G \in \mathbb{R}^{C \times C}$ denotes the weights of the GCN, and ReLU is the activation function. 
The features $\hat{\X}_G $ are acquired by the vertex-wise interaction (multiplication with $(\I - \A)$) and channel-wise interaction (multiplication with $\W_G$). 
We set the same number of output channels as the input to keep consistency, allowing the module to be compatible with the subsequent process.

The main difference between previous GCNs \cite{kipf2016semi,luo2020every} and ours is that, instead of a hand-crafted graph in \cite{kipf2016semi} or a graph with available connectivities \cite{luo2020every}, we randomly initialize the graph $\A$ and then iteratively learn BOTH the connectivity and edge weights by a linear layer during the training, which adaptively captures the implicit relations among facial components.
Besides, we add a residual connection to reserve the input features of vertices $\X_G$. 
The finally reasoned relations are acquired with the information propagation across all vertices based on the learned graph. After graph reasoning, the semantic information is greatly enhanced across different vertices.

\subsection{Graph Reprojection}
\label{subsec:reprojection}

Having learned vertex features that capture the semantic information of each facial component, we project vertex features in the graph domain back to the original pixel domain for face parsing. 
That is, we aim to compute a matrix $\P:\hat{\X}_G \mapsto \X_P$ that reprojects vertex features $\hat{\X}_G$ to pixel features $\X_P$. 

In particular, we leverage vertex features $\X_{G}$ prior to reasoning and edge-masked pixel features $\X_e$ to construct the projection matrix $\P$, which models the correlation between vertices and pixels. 
Specifically, we take the inner product of $\X_{G}$ and $\X_e$ to capture the similarity between vertices and each pixel. We then apply a softmax function for normalization. Formally, the projection matrix takes the form:
\begin{equation}
 \P = \text{softmax}\left(\X_{G} \cdot \X_e^{\top}\right),
 \label{eq:projection_matrix}
\end{equation}
where $\P \in \mathbb{R}^{KN_c \times H_1W_1}$. 

Having acquired the projection matrix $\P$ modeling the correlation between vertices and pixels, we obtain the final refined pixel-level feature map $\X_P$ by taking the product of the projection matrix $\P$ and refined vertex features $\hat{\X}_G$, \ie,
\begin{equation}
    \X_P = \P^{\top} \cdot \hat{\X}_G.
\end{equation}

After reprojection, in order to take advantage of the original feature map $\X_0$ that captures features of pixel-level texture information, we add $\X_0$ to the refined feature map $\X_P$ that captures edge-aware component-level semantic information by element-wise summation, leading to {\it multi-level} feature representations. 
Followed by a $1 \times 1$ convolution, our model predicts the final parsing result $\Y$ as
\begin{equation}
    \Y = \text{conv}(\X_0 + \P^{\top} \cdot \hat{\X}_G),
\end{equation}
where $\text{conv}(\cdot)$ denotes the $1 \times 1$ convolution.

%% file: 3_training.tex
Having presented our model, we propose a boundary-attention loss and a discriminative loss tailored to the training of our model in Section~\ref{subsec:loss}. Further, we provide detailed analysis and discussion in Section~\ref{subsec:analysis}.

\subsection{The Loss Function}
\label{subsec:loss}

In addition to the commonly adopted parsing loss, the training objective of AGRNet further aims to promote the segmentation accuracy along the boundary as well as learning discriminative vertex features. 
Hence, we propose a boundary-attention loss and a discriminative loss respectively, which are detailed below.  

\subsubsection{The Proposed Boundary-Attention Loss}
To improve the segmentation results along the boundary, we introduce the boundary-attention loss (BA-Loss) inspired by \cite{liu2020new} as in our previous work \cite{te2020edge}.
The BA-loss measures the loss of the predicted parsing labels compared to the ground truth {\it only} at edge pixels, thus strengthening the model capacity for critical edge pixels that are challenging to distinguish.
Mathematically, the BA-loss is formulated as 
\begin{equation}
    \mathcal{L}_{\text{BA}} = \sum_{i=1}^{HW}\sum_{j=1}^{N_c}\left[ e_i = 1 \right] y_{ij}\log{p_{ij}},
    \label{eq:BA-loss}
\end{equation}
where $i$ is the pixel index, $j$ is the class index and $N_c$ is the number of categories. $e_i$ is a binary scalar to indicate an edge pixel ($e_i=1$), $y_{ij}$ denotes the ground truth label of face parsing, and $p_{ij}$ denotes the predicted parsing label. 
$\left[ \cdot \right]$ is the Iverson bracket, which denotes a number that is $1$ if the condition in the bracket is satisfied, and $0$ otherwise. 

\subsubsection{The Proposed Discriminative Loss}
In semantic segmentation, pixels belonging to different categories should be distant from each other in the feature space, while pixels belonging to the same category should be as similar as possible. 
In our model, while the parsing map provides hints in selecting semantic vertices, the similarity among different vertices is however overlooked. 
The separation of vertices is also critical in pushing pixels to its corresponding component. 

Motivated by \cite{de2017semantic} \cite{huang2020ccnet} \cite{luo2021category} \cite{luo2019significance}, we propose a discriminative loss that penalizes small feature distances between vertices representing different components and encourages multiple vertices corresponding to the same component to be more diverse.
Specifically, let $\mathbf{x_1}$ and $\mathbf{x_2}$ denote the features of two semantic vertices respectively, we formulate the penalty function as
\begin{equation}
\label{eq:discriminative_loss}
    \phi(\mathbf{x_1}, \mathbf{x_2}) = 
    \left\{
    \begin{aligned}
       &(\delta - ||\mathbf{x_1} - \mathbf{x_2}||_2)^2,  &||\mathbf{x_1}-\mathbf{x_2}||_2 < \delta
        \\
        &0, &||\mathbf{x_1}-\mathbf{x_2}||_2 \geq \delta
    \end{aligned}
  \right.
  ,
\end{equation}
where $\delta$ is a pre-defined threshold to control the threshold of the feature distance between two semantic vertices.
If the $l_2$ distance between the two features exceeds $\delta$, the function does not impose any penalty. 
Otherwise, the penalty is a quadratic function that takes a larger value for the smaller distance.

Considering all the vertices with cardinality $KN_c$, we formulate the complete discriminative loss $\mathcal{L}_{\text{dis}}$ as follows:
\begin{equation}
    \mathcal{L}_{\text{dis}} = \frac{1}{KN_c(KN_c-1)}\sum_{i=1}^{KN_c}{\sum_{j=1, j \neq i}^{KN_c}{\phi(\mathbf{x}_i, \mathbf{x}_j)}}.
\end{equation}

\subsubsection{The Total Loss}
In addition to the above two loss functions, we have three more losses: 
1) the prediction loss of the raw parsing map for adaptive graph projection $\mathcal{L}_{\text{raw}}$, which takes the cross entropy between the predicted raw parsing map and the ground truth parsing map; 
2) the final parsing loss $\mathcal{L}_{\text{final}}$, which takes the cross entropy between each predicted label and the ground truth label; 
3) the edge prediction loss $\mathcal{L}_{\text{edge}}$, which measures the estimation error of the image edges. 
The three loss functions take the following forms:
\begin{equation*}
\begin{aligned}
    \mathcal{L}_{\text{raw}} =& -\frac{1}{HW}\sum_{i=1}^{HW}{\sum_{j=1}^{N_c}{y_{ij}\log(\text{pred}^{raw}_{ij})}};\\
    \mathcal{L}_{\text{final}} =& -\frac{1}{HW}\sum_{i=1}^{HW}{\sum_{j=1}^{N_c}{y_{ij}\log(\text{pred}^{final}_{ij})}};\\
    \mathcal{L}_{\text{edge}} = &-\frac{1}{HW}\sum_{i=1}^{HW}{e_{i}\log(\text{pred}^{edge}_{i})} \\ &+ (1 - e_{i})\log(1 - \text{pred}^{edge}_{i}). 
\end{aligned}
\end{equation*}

The total loss function is then defined as follows:
\begin{equation}
    \mathcal{L} = \lambda_1\mathcal{L}_{\text{raw}} + \lambda_2\mathcal{L}_{\text{edge}} + \lambda_3\mathcal{L}_{\text{BA}} + \lambda_4\mathcal{L}_{\text{final}} + \lambda_5\mathcal{L}_{\text{dis}},
    \label{eq:loss_total}
\end{equation}
where $\{\lambda_{i}\}_{i=1}^5$ are hyper-parameters to strike a balance among different loss functions.  

Finally, we provide a detailed training pipeline in Algorithm \ref{alg:training}. 

\begin{algorithm}
\caption{Training for AGRNet}
\label{alg:training}
\linespread{1}
\selectfont
\DontPrintSemicolon
  \KwInput{Image set $\{\mathbf{I_1},...,\mathbf{I_n}\}$, Edge set $\{\mathbf{E_1},...,\mathbf{E_n}\}$, Label set $\{\mathbf{Y_1},...,\mathbf{Y_n}\}$, \\
  ~~~~~~~~~Hyper parameters $\lambda_1,...,\lambda_5$, Number of categories $N_c$}
  \KwOutput{Parsing Map $\Y$, Model parameters $\Phi$}
  \While{not converge}
  {
    Sample $\mathbf{I_i}$, $\mathbf{E_i}$, $\mathbf{Y_i}$ from $\{\mathbf{I_1},...,\mathbf{I_n}\}, \{\mathbf{E_1},...,\mathbf{E_n}\}, \{\mathbf{Y_1},...,\mathbf{Y_n}\}$\\
    Acquire a low-level/high-level feature map $\mathbf{X_1}, \mathbf{X_2}$ = $Resnet$($\mathbf{I_i}$)\\
    $\mathbf{X_0}$ = $Fuse([\mathbf{X_1}, \mathbf{X_2}])$ as in (1)\\
    Learn an edge map $\mathbf{E}$ = $Edge Module$($\mathbf{X_0}$)\\
    Predict a preliminary parsing map $\mathbf{Z_0}$ = $Predict$($\mathbf{X_0}$) \\
    
    \tcp{Adaptive Graph Projection}
    $\X_{e} = \X_0 \circ \mathbf{E}$ \\
    $\X_{ne} = \X_0 \circ (\mathbf{1} - \mathbf{E})$ \\
    $\X_{G} = \X_{ne}\{\parallel_{i=1}^{N_c}\text{topk}(\mathbf{Z_0}[:,i])\}$ \\
    
    \tcp{Graph Learning and Reasoning}
        $\hat{\X}_G = \text{ReLU}\left[(\I - \A)\X_G\W_G\right]$ \\
        
    \tcp{Graph Reprojection}
    Compute the projection matrix $\mathbf{P} = \text{softmax}\left(\X_{G} \cdot \X_e^{\top}\right)$ \\
    Obtain the final refined pixel-level feature map via graph reprojection $\X_P = \mathbf{P}^{\top} \cdot \hat{\X}_G$ \\
    Predict the final parsing result $\Y = \text{conv}(\X_0 + \X_P)$ 
    
    \tcp{Loss Aggregation}
    $\mathcal{L} = \lambda_1\mathcal{L}_{\text{raw}} + \lambda_2\mathcal{L}_{\text{edge}} + \lambda_3\mathcal{L}_{\text{BA}} + \lambda_4\mathcal{L}_{\text{final}} + \lambda_5\mathcal{L}_{\text{dis}}$ as in (11) \\
    Update network parameters with the SGD optimizer and loss $\mathcal{L}$\\
  }
\end{algorithm}

\begin{table*}[!ht]
\centering
\caption{Ablation study on the LaPa dataset. We conduct comparison experiments from multiple aspects, including composition of modules, loss functions, vertex numbers, FLOPs and parameters.}
\label{table:ablation}
    \begin{subtable}{\textwidth}
    \centering
    \begin{tabular}{c|ccccc|ccc}
    \toprule
    Model & ResNet & Spatial & Adaptive & Edge & Graph & Mean F1 (\%) & Mean IoU (\%) & Mean Accuracy (\%) \\ 
    \midrule
    1 & \checkmark & & & & & 90.9 & 85.1 & 90.2 \\ 
    2 & \checkmark & \checkmark & &  &  & 91.1 & 85.7 & 90.8 \\ 
    3 & \checkmark & & \checkmark & & & 91.4 & 86.0 & 91.3  \\ 
    4 & \checkmark & & \checkmark & & \checkmark & 91.9 & 86.3 & 91.7 \\ 
    5 & \checkmark & & \checkmark & \checkmark & & 91.6 & 85.6 & 92.1 \\ 
    6 & \checkmark & & \checkmark & \checkmark & \checkmark & \textbf{92.3} & \textbf{87.0} & \textbf{92.7} \\ 
    \bottomrule
    \end{tabular}
    \vspace{0.5em}
    \caption{On composition of modules}
    \end{subtable}
\bigskip
    \begin{subtable}{\textwidth}
    \centering
    \begin{tabular}{c|ccc|ccc}
    \toprule
    Model & Cross Entropy & Discriminative & Boundary-Attention & Mean F1 (\%) & Mean IoU (\%)& Mean Accuracy (\%)\\ 
    \midrule
    1 & \checkmark & & & 90.9 & 85.1 & 90.2 \\ 
    2 & \checkmark & \checkmark & & 91.4 & 86.1 & 92.5 \\ 
    3 & \checkmark & & \checkmark & 91.8 & 86.3 & 91.8 \\ 
    4 & \checkmark & \checkmark & \checkmark & \textbf{92.3} & \textbf{87.0} & \textbf{92.7} \\ 
    \bottomrule
    \end{tabular}
    \vspace{0.5em}
    \caption{On loss functions}
    \label{table:ablation_loss}
    \end{subtable}
\bigskip
    \begin{subtable}{0.4\textwidth}
    \begin{tabular}{c|c|c|c|c}
    \toprule
    \begin{minipage}{2cm} \centering  \vspace{1mm} \end{minipage} & \multicolumn{4}{c}{Top K} \\
    \hline
    Metric (\%) &  top 1 & top 2 & top 4 & top 8 \\ 
    \hline
    Mean F1 & 91.6 & 91.9 & \textbf{92.3} & 91.1\\ 
    Mean IoU   & 86.0 & 86.3 & \textbf{87.0} & 86.2\\ 
    Mean Accuracy   & 92.0 & \textbf{93.5} & 92.7 & 92.3\\ 
    \bottomrule
    \end{tabular}
    \vspace{0.5em}
    \caption{On vertex numbers}
    \label{table:ablation_anchors}
    \end{subtable}
    \begin{subtable}{0.4\textwidth}
    \begin{tabular}{c|cc}
    \toprule
                      & FLOPs (G) & Parameters (M) \\
    \hline
    ResNet           & 6.652    & 14.10         \\
    Edge Perceiving  & 0.249    & 0.03         \\
    Graph Projection & 0.471    & 0.02         \\
    Graph Reasoning  & 0.037    & \textless0.01        \\
    \bottomrule
    \end{tabular}
    \vspace{0.5em}
    \caption{On FLOPs and parameters}
    \label{table:ablation_complexity}
    \end{subtable}
\end{table*}

\subsection{Difference from Relevant Models} 
\label{subsec:analysis}
While the proposed graph projection is inspired by non-local modules and the entire model belongs to the family of graph-based methods, we analyze the prominent differences between previous works and our method. 

\textbf{Comparison with non-local modules.}
Typically, a traditional non-local module models {\it pixel-wise} correlations based on pixel-wise feature similarities, while neglecting the high-order relationship among regions. 
In contrast, we explicitly exploit the correlation among distinct components via the proposed pixel-to-vertex projection and graph reasoning over vertices. 
Each vertex embeds certain facial component, which models the most prominent characteristics towards semantics.
We further learn and reason over the relations between vertices by graph convolution, which captures high-order semantic relations among different facial components.

Also, the computation complexity of non-local modules is expensive in general \cite{wang2018non}.  
The proposed edge-aware adaptive pooling addresses this issue by extracting a few significant vertices instead of redundant query points, which reduces the attention map size from $O(N^2)$ to $O(NK) $, where $K \ll N$. 
Further, we separate and assign different weights to edge and non-edge pixels via the learned edge map, which emphasizes on challenging edge pixels, thus improving the parsing quality.

\textbf{Comparison with graph-based models.}
In comparison with other graph-based models, such as \cite{chen2019graph,li2018beyond,te2020edge}, we significantly improve the graph projection process by introducing semantics in an adaptive and differentiable manner. 
It is worth noting that, the vertex selection is critical in the graph construction. 
In previous works, each vertex is simply represented by a weighted sum of image pixels or acquired by spatial pooling, which may result in ambiguity in understanding vertices. 
Besides, given different inputs of feature maps, the pixel-wise features often vary greatly but the projection matrix is fixed after training. 
In contrast, we incorporate explicit semantics into the projection process to extract robust vertices and construct the projection matrix {\it on-the-fly} on the basis of the similarity between vertices and each pixel in the feature space.

Furthermore, in comparison with hand-crafted graph models \cite{gong2019graphonomy, he2020grapy}, which are broadly used in human parsing, our model learns the implicit graph and the correlation between facial components. Different from the human structure which implies the graph connectivity, there is no conventional topology predefined on human faces. Therefore, we propose the dynamically learned graph reasoning module to infer the underlying graph adjacency matrix instead of the hand-crafted human structure in Graphonomy \cite{gong2019graphonomy}. 

\subsection{Theoretical analysis}

In particular, our method has addressed existing major challenges of face parsing from the following two crucial aspects.

Firstly, {\it the correlation between different facial components} is critical in face parsing. However, existing methods, including face-based and region-based parsing, overlook such correlation among different parts. Face-based parsing may neglect the scale discrepancy in different facial components and region-based parsing does not exploit the inter-region relationship and long-range dependencies, while it would definitely enhance the parsing performance to explore such correlations. 
Similar theory has been validated on the task of human parsing \cite{gong2019graphonomy}. 
Further, facial structural variance is much less than human structure, leading to more stable structural correlations for effective learning. 
Graph representation is one of the most effective ways to model such correlations. Hence, we introduce a graph-based model for face representation and employ graph convolutional networks to construct and reason the component-wise relationship.

Secondly, it remains a challenge to segment pixels around the edges between components, since edges tend to be ambiguous in real-world scenarios. This issue is more severe in face parsing, as edge pixels cover a higher proportion in face images than that for other tasks such as scene parsing.
Therefore, improving the segmentation accuracy of edge pixels is one effective way to enhance the performance of face parsing.
In this paper, we achieve accurate segmentation along the edges from two aspects: 1) we incorporate edge attention in the pixel-to-vertex projection, which assigns larger weights to the features of edge pixels during the feature aggregation; 2) we design a boundary-attention loss to reinforce edge pixels, which measures the error of the predicted parsing map only at edge pixels.

In face of the above two challenges and based on the principles to address them, we explicitly model the component-wise correlations on a graph abstraction with edge pixels taken into consideration, where each vertex describes a facial component and each link captures the correlation between a pair of components. 
To learn such a graph representation, we propose adaptive graph representation learning and reasoning over facial components, which learns and reasons over non-local components to capture long range dependencies with the boundary information between components leveraged. 
To obtain representative vertices, we adaptively select pixels with high responses to a distinct facial component as a graph vertex. Since facial components are unknown in the beginning, we employ a predicted parsing map as the initial condition, and propose a discriminative loss to enhance the discrimination between vertices, leading to more representative and distinct vertices with strong semantics. 
Further, we integrate the component-wise correlation into the feature representation of edge pixels, so that it contains not only the local contextual information, but also the correlation with other components, thus enhancing the semantics.

The above theoretical analysis provides the interpretability of our method for deep-learning based face analysis, which can be summarized as follows:
we propose an effective network to address two great challenges in face parsing---insufficient use of component-wise correlations and ambiguities of edge pixels, and design several losses to guide the network to adaptively learn the intrinsic features of face images from data.

%% file: 4_results.tex
To validate the proposed AGRNet, we conduct extensive experiments on face parsing as well as on human parsing for generalizability.

\subsection{Datasets and Metrics}

\begin{table*}[htbp]
\centering
\caption{Comparison with the state-of-the-art methods on face parsing datasets (in F1 score).}
\label{table:comparison}
    \begin{subtable}{\textwidth}
    \centering
    \begin{tabular}{c|cccccccc|c}
    \toprule
    Methods & Skin & Nose & U-lip & I-mouth & L-lip & Eyes & Brows & Mouth & Overall \\ 
    \midrule
    
    Liu \et \cite{liu2017face} & 92.1 & 93.0 & 74.3 & 79.2 & 81.7 & 86.8 & 77.0 & 89.1 & 88.6 \\ 
    Lin \et \cite{lin2019face} & 94.5 & 95.6 & 79.6 & 86.7 & 89.8 & 89.6 & 83.1 & 95.0 & 92.7 \\ 
    Wei \et \cite{wei2019accurate} & \textbf{95.6} & 95.2 & 80.0 & 86.7 & 86.4 & 89.0 & 82.6 & 93.6 & 91.7 \\ 
    Yin \et \cite{yin2020end} & - & \textbf{96.3} & 82.4 & 85.6 & 86.6 & 89.5 & 84.8 & 92.8 & 91.0 \\ 
    Liu \et \cite{liu2020new} & 94.9 & 95.8 & \textbf{83.7} & 89.1 & \textbf{91.4} & 89.8 & 83.5 & 96.1 & 93.1 \\
    Te \et\cite{te2020edge} & 94.6 & 96.1 & 83.6 & 89.8 & 91.0 & \textbf{90.2} & 84.9 & 95.5 & 93.2 \\
    \midrule
    Ours & 95.1 & 95.9 & 83.2 & \textbf{90.0} & 90.9 & 90.1 & \textbf{85.0} & \textbf{96.2} & \textbf{93.2} \\
    \bottomrule
    \end{tabular}
    \vspace{0.5em}
    \caption{The Helen dataset}
    \end{subtable}

    \begin{subtable}{\textwidth}
    \centering
    \begin{tabular}{c|cccccccccc|c}
    \toprule
    Methods & Skin & Hair & L-Eye & R-Eye & U-lip & I-mouth & L-lip & Nose & L-Brow & R-Brow & Mean \\ 
    \midrule
    Zhao \et \cite{zhao2017pyramid} & 93.5 & 94.1 & 86.3 & 86.0 & 83.6 & 86.9 & 84.7 & 94.8 & 86.8 & 86.9  & 88.4 \\
    Liu \et \cite{liu2020new}       & 97.2 & 96.3 & 88.1 & 88.0 & 84.4 & 87.6 & 85.7 & 95.5 & 87.7 & 87.6 & 89.8 \\
    Te \et \cite{te2020edge}         & 97.3 & 96.2 & 89.5 & 90.0 & 88.1 & 90.0 & 89.0 & 97.1 & 86.5 & 87.0 & 91.1 \\
    Luo \et \cite{luo2020ehanet} & 95.8 & 94.3 & 87.0 & 89.1 & 85.3 & 85.6 & 88.8 & 94.3 & 85.9 & 86.1 & 89.2 \\
    Wei \et \cite{wei2019accurate} & 96.1 & 95.1 & 88.9 & 87.5 & 83.1 & 89.2 & 83.8 & 96.1 & 86.0 & 87.8 & 89.4 \\
    \midrule
    Ours & \textbf{97.7} & \textbf{96.5} & \textbf{91.6} & \textbf{91.1} & \textbf{88.5} & \textbf{90.7} & \textbf{90.1} & \textbf{97.3} & \textbf{89.9} & \textbf{90.0} & \textbf{92.3} \\
    \bottomrule
    \end{tabular}
    \vspace{0.5em}
    \caption{The LaPa dataset}
    \end{subtable}
    
    \begin{subtable}{\textwidth}
    \centering
    \begin{tabular}{c|ccccccccc|c}
    \toprule
    \multirow{2}{*}{Methods}  & Face    & Nose  & Glasses & L-Eye & R-Eye & L-Brow  & R-Brow   & L-Ear & R-Ear & \multirow{2}{*}{Mean} \\
                              & I-Mouth & U-Lip & L-Lip   & Hair  & Hat   & Earring & Necklace & Neck  & Cloth &                       \\
    \midrule
    \multirow{2}{*}{Zhao \et \cite{zhao2017pyramid}}   & 94.8    & 90.3  & 75.8    & 79.9  & 80.1  & 77.3    & 78.0       & 75.6  & 73.1  & \multirow{2}{*}{76.2} \\
                              & 89.8    & 87.1  & 88.8    & 90.4  & 58.2  & 65.7    & 19.4     & 82.7  & 64.2  &                       \\
    \midrule
    \multirow{2}{*}{Lee \et\cite{CelebAMask-HQ}} & 95.5    & 85.6  & \textbf{92.9}    & 84.3  & 85.2  & 81.4    & 81.2     & 84.9  & 83.1  & \multirow{2}{*}{80.3} \\
                              & 63.4    & 88.9  & 90.1    & 86.6  & \textbf{91.3}  & 63.2    & 26.1     & \textbf{92.8}  & 68.3  &                       \\
    \midrule
    \multirow{2}{*}{Luo \et\cite{luo2020ehanet}} & 96.0 & 93.7 & 90.6 & 86.2 & 86.5 & 83.2 & 83.1 & 86.5 & 84.1  & \multirow{2}{*}{84.0} \\
                              & 93.8 & 88.6 & 90.3 & 93.9 & 85.9 & 67.8 & 30.1 & 88.8 & 83.5  &                       \\
    \midrule
    \multirow{2}{*}{Wei \et \cite{wei2019accurate}} & 96.4    & 91.9    & 89.5    & 87.1  & 85.0  & 80.8   & 82.5     & 84.1    & 83.3  & \multirow{2}{*}{82.1} \\
                              & 90.6      & 87.9  & 91.0    & 91.1  & 83.9  & 65.4    & 17.8     & 88.1  & 80.6  &    \\  
    \midrule
    \multirow{2}{*}{Te \et\cite{te2020edge}}     & 96.2    & \textbf{94}    & 92.3    & 88.6  & 88.7  & \textbf{85.7}    & 85.2     & 88    & 85.7  & \multirow{2}{*}{85.1} \\
                              & \textbf{95.0}      & 88.9  & \textbf{91.2}    & \textbf{94.9}  & 87.6  & 68.3    & \textbf{27.6}     & 89.4  & \textbf{85.3}  &    \\  
    \midrule
    \multirow{2}{*}{Ours}     & \textbf{96.5}    & 93.9    & 91.8    & \textbf{88.7}  & \textbf{89.1}  & 85.5    & \textbf{85.6}     & \textbf{88.1}    & \textbf{88.7}  & \multirow{2}{*}{\textbf{85.5}} \\
                              & 92.0      & \textbf{89.1}  & 91.1    & 95.2  & 87.2  & \textbf{69.6}    & \textbf{32.8}     & 89.9  & 84.9  &    \\  
    \bottomrule
    \end{tabular}
    \vspace{0.5em}
    \caption{The CelebAMask-HQ dataset}
    \end{subtable}
\end{table*}


\textbf{Face Parsing} The Helen dataset \cite{le2012interactive} includes 2,330 images with 11 categories: background, skin, left/right brow, left/right eye, upper/lower lip, inner mouth and hair. Specifically, we keep the same training/validation/test protocol as in \cite{le2012interactive}. The number of the training, validation and test samples are 2,000, 230 and 100, respectively. 
The LaPa dataset \cite{liu2020new} is a large-scale face parsing dataset, consisting of more than 22,000 facial images with abundant variations in expression, pose and occlusion, and each image is provided with an 11-category pixel-level label map and 106-point landmarks. 
The CelebAMask-HQ dataset \cite{CelebAMask-HQ} is composed of 24,183 training images, 2,993 validation images and 2,824 test images. The number of categories in CelebAMask-HQ is 19. In addition to facial components, the accessories such as eyeglass, earring, necklace, neck, and cloth are also annotated.

\textbf{Human Parsing} 
The LIP dataset \cite{gong2017look} is a large-scale dataset focusing on semantic understanding of person, which contains 50,000 images with elaborated pixel-wise annotations of 19 semantic human part labels as well as 2D human poses with 16 key points. The images collected from the real-world scenarios contain human appearing with challenging poses and views, serious occlusions, various appearances and low-resolutions.

\textbf{Metrics}: We employ three evaluation metrics to measure the performance of our model: pixel accuracy, intersection over union (IoU) and F1 score. The mean value is calculated by the average of total categories.
Directly employing the pixel accuracy metric ignores the scale variance amid facial components, while the mean IoU and F1 score are better for evaluation. 
To keep consistent with the previous methods, we report the overall F1 score on the Helen dataset, which is computed over the merged facial components: brows (left + right), eyes (left + right), nose, mouth (upper lip + lower lip + inner mouth).

\subsection{Implementation Details}


During the training, we employ the random rotation and scale augmentation. The rotation angle is randomly selected from $(-30^\circ, 30^\circ)$ and the scale factor is randomly selected from $(0.75, 1.25)$. The ground truth of the edge mask is extracted according to the semantic label map. If the label of a pixel is different from any of its four neighbors, it is regarded as an edge pixel. For the Helen dataset, we pre-process the face image by face alignment as in other works. For the LaPa and CelebAMask-HQ datasets, we directly utilize the original images without any pre-processing.

We take the ResNet-101\cite{he2016deep} as the backbone and extract the output of conv\_2 and conv\_5 layers as the low-level and high-level feature maps for multi-scale representations. 
To reduce the information loss in the spatial space, we utilize the dilation convolution in the last two blocks and the output size of the final feature map is 1/8 of the input image. 
To fully exploit the global information in high-level features, we employ a spatial pyramid pooling operation \cite{zhao2017pyramid} to learn multi-scale contextual information. 
In the edge prediction branch, we concatenate the output of conv\_2, conv\_3, conv\_4 layers and apply a 1 $\times$ 1 convolution to predict the edge map. 
In the differentiable graph projection, we set top-\textbf{4} pixels as representative vertices for each face component, and thus the number of graph vertices is 4 times of the category number. 
We choose the hyper-parameters in \eqref{eq:loss_total} as $\lambda_1$ = 1, $\lambda_2$ = 1, $\lambda_3$ = 1, $\lambda_4$ = 0.5, $\lambda_5$ = 0.1 by grid search and according to prior knowledge\footnote{The first four losses in \eqref{eq:loss_total} are in the form of logarithm, while the last loss is in the form of the Euclidean distance. Hence, the weighting parameters of the first four losses should be larger than the parameter of the last loss.} of the scale of each loss for the initial setting of grid search.
In the proposed discriminative loss in \eqref{eq:discriminative_loss}, we normalize the vertex features and set $\delta=1$.

All the experiments are implemented with 4 NVIDIA RTX 2080Ti GPUs. Stochastic Gradient Descent (SGD) is employed for optimization. We initialize the network with a pre-trained model on ImageNet. The input size is $473 \times 473$ and the batch size is set to 8. The learning rate starts at 0.001 with the weight decay of 0.0005. 
The batch normalization is implemented with In-Place Activated Batch Norm \cite{rotabulo2017place}.  For fair comparisons, these settings are adopted for all the compared methods.

\begin{figure*}[tbp]
    \centering
    \includegraphics[width=0.8\textwidth]{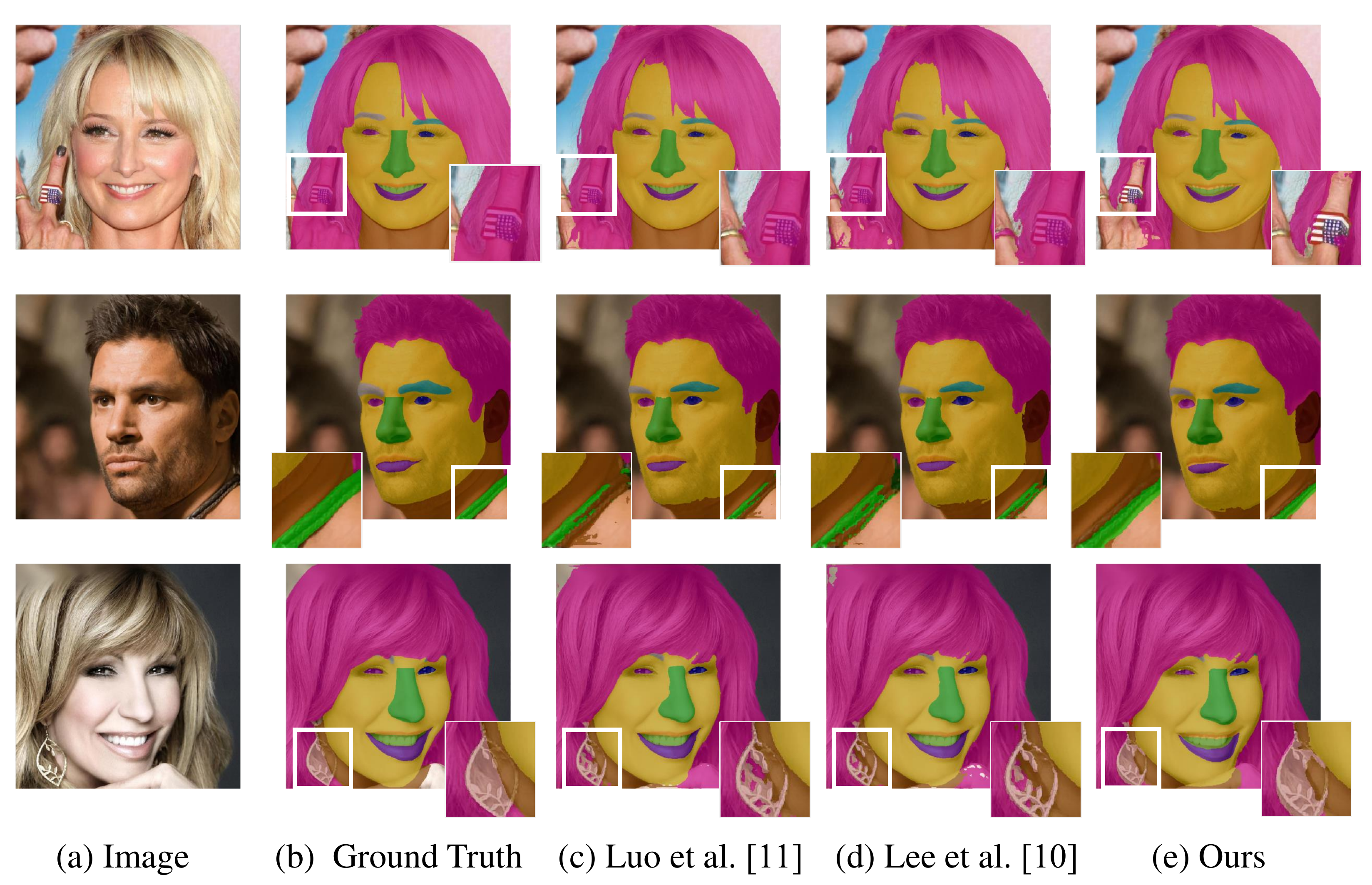}
    \caption{Visualization of parsing results from different methods on the CelebAMask-HQ dataset.}
    \label{fig:comparison}
\end{figure*}

\subsection{Face Parsing}

We perform comprehensive experiments on the commonly used three face parsing datasets and present the experimental results as follows.

\subsubsection{Ablation study}

The ablation study is conducted from multiple aspects on the LaPa dataset.

\textbf{On composition of modules.} 
We evaluate the improvement brought by different modules in AGRNet. 
Specifically, we remove some components and train the model from scratch under the same initialization. The quantitative results are reported in Table~\ref{table:ablation}-(a), where {\it ResNet} denotes the baseline model trained with the network consisting of the backbone and the pyramid spatial pooling. It achieves 90.9\% in Mean F1 score.
{\it Spatial} and {\it Adaptive} refer to the respective schemes of graph construction. 
{\it Spatial} denotes uniform spatial pooling as in the previous work \cite{te2020edge} and {\it Adaptive} denotes the proposed adaptive pooling method in this paper, which improves the F1 score of Model 1 ({\it ResNet}) by 0.2\% and 0.5\%, respectively.
This result validates the effectiveness of the proposed adaptive pixel-to-vertex projection. 
{\it Edge} and {\it Graph} represent the edge module and graph reasoning module, respectively. 
Compared with Model 3, the edge module achieves improvement of 0.2\% in Mean F1 score, while the {\it Graph} module improves the Mean F1 score by 0.5\%. 
This demonstrates the superiority of the proposed graph reasoning. 
By employing all the modules, we achieve the best performance of 92.3\% in Mean F1 score.

\textbf{On loss functions.} We evaluate the effectiveness of different loss functions described in Section~\ref{subsec:loss} and show the comparison results in Table~\ref{table:ablation}-(b). We observe that the Discriminative loss leads to 0.5\% of improvement and the Boundary-Attention loss brings 0.9\% of improvement in Mean F1 score compared with the traditional cross entropy loss. The best performance is obtained by utilizing all of these loss functions.

\textbf{On vertex numbers.} Table~\ref{table:ablation}-(c) presents the performance with respect to different number of vertices. Our model achieves the best mIoU and overall F1 under the top 4 setting. We suppose that too many vertices result in redundant noisy vertices whereas fewer vertices fail to represent enough semantic information.

\textbf{On FLOPs and parameters.} 
We compute the complexity of different modules in FLOPs and Parameter size in Table~\ref{table:ablation}-(d). 
All the numbers are computed with the input image size of 473 $\times$ 473 when the batch size is set to 1.
ResNet refers to the backbone network, which has 6.652GFLOPs and 14.10M of parameters. The Edge Perceiving module brings only about 0.25GFLOPs and 0.03M of parameters. 
The Graph Projection module further leads to only 0.471G and 0.02M of increase in FLOPs and parameters, respectively. After integrating the Graph Reasoning module, the network has 7.411GFLOPs and 14.15M of parameters. 
We see that our model improves the F1 score by 1.4\% at the cost of mere 0.8GFLOPs and 0.05M of parameters compared with the ResNet backbone.  

\begin{figure}[htbp]
    \centering
    \includegraphics[width=0.45\textwidth]{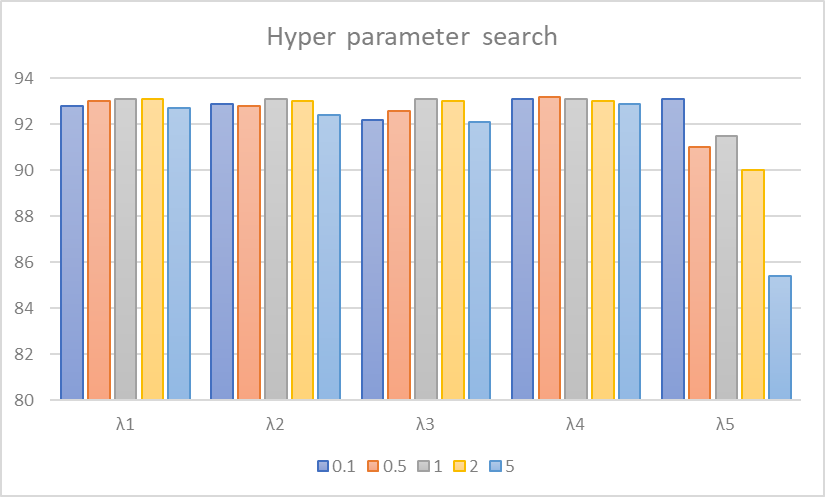}
    \caption{Visualization of the hyper-parameter search.}
    \label{fig:hyper}
\end{figure}

\textbf{On hyper-parameters.}

In the experiments, we set the parameters $\{\lambda_i\}_{i=1}^4$ in \eqref{eq:loss_total} based on the setting of the LaPa work \cite{liu2020new}, which achieves convincing performance on several face parsing datasets. As to the weighting parameter of the discriminative loss---$\lambda_5$, we assign its value according to the approximate proportion of loss scale.
For more rigorous parameter settings, we additionally adopt the traditional way of performing hyper-parameter optimization---{\it grid search} or a parameter sweep, which is an exhaustive searching through a manually specified subset of the hyper-parameter space of a learning algorithm. 
In our case, the grid search algorithm is guided by the evaluation metric of Overall F1 score on the test set of the {\it Helen} dataset. 

Specifically, we train our model under a finite set of reasonable settings of hyper-parameters $\{\lambda_i\}_{i=1}^5$. We change a single parameter at a time, and report the respective results in Table \ref{table:hyper}. 
We see that a choice of these parameters $\lambda_1 =1,\lambda_2 =1,\lambda_3 =1,\lambda_4 =0.5,\lambda_5=0.1$ is empirically optimal in this case, which is our setting. Thus, in general, our choice of the parameters is empirically optimal.

Further, we illustrate the grid search in Fig.~\ref{fig:hyper}. 
We observe that the results are insensitive to $\lambda_1, \lambda_2,\lambda_3,\lambda_4$, while they show sensitivity to $\lambda_5$---the weighting parameter of the discriminative loss. This is mainly because the discriminative loss has less contribution to the final segmentation result than the cross entropy loss. 
Therefore, assigning a large weight to the discriminative loss will weaken the supervision of the ground truth parsing map and thus influence the overall performance.

\begin{table}[t]
\centering
\caption{The parsing results under different hyper-parameters (measured by overall F1).}
\label{table:hyper}
\begin{tabular}{c|cccccccc|c}
\toprule
Parameter & =0.1 & =0.5 & =1 & =2 & =5 \\ 
\midrule

$\lambda_1$ & 92.8 & 93.0 & \textbf{93.1} & 93.1 & 92.7 \\
$\lambda_2$ & 92.9 & 92.8 & \textbf{93.1} & 93.0 & 92.4 \\
$\lambda_3$ & 92.2 & 92.6 & \textbf{93.1} & 93.0 & 92.1 \\
$\lambda_4$ & 93.1 & \textbf{93.2} & 93.1 & 93.0 & 92.9 \\
$\lambda_5$ & \textbf{93.1} & 92.0 & 91.5 & 90.0 & 85.4 \\
\bottomrule
\end{tabular}
\end{table}



\begin{table*}[t]
\centering
\caption{The performance over domain gaps (measured in F1 score).}
\label{table:cross}
\begin{tabular}{c|c|cccccccc|c}
\toprule
Train & Test & Skin & Nose & U-lip & I-mouth & L-lip & Eyes & Brows & Mouth & Overall \\ 
\midrule

Helen-Train & Helen-Test & 95.0 & 96.0 & 83.1 & 90.0 & 90.7 & 89.9 & 85.1 & 96.2 & 93.1 \\
\midrule
CelebAMask-HQ & Helen-Test  & 93.2 & 94.7 & 82.7 & 88.5 & 90.1 & 89.1 & 79.6 & 92.4 & 91.8  \\
\midrule
Helen-Primary & Helen-Reference & 92.6 & 92.0 & 74.1 & 91.0 & 89.2 & 81.2 & 83.1 & 95.6 & 90.9  \\
Helen-Reference & Helen-Primary & 61.3 & 34.8 & 4.6 & 0 & 45.1 & 69.4 & 30.7 & 33.2 & 42.9  \\
\bottomrule
\end{tabular}
\end{table*}

\begin{table*}[htbp]
\centering
\caption{Experimental comparison on the LIP dataset for human parsing (in IoU score).}
\label{table:LIP}
\resizebox{\textwidth}{!}{
\begin{tabular}{c|cccccccccccccccccccc|c}
\toprule 
     & background & hat   & hair  & glove & glasses & u-cloth & dress & coat  & socks & pants & j-suits & scarf & skirt & face & l-arm & r-arm & l-leg & r-leg & l-shoe & r-shoe & mean \\
\midrule 
SegNet~\cite{badrinarayanan2017segnet} &70.62&26.60&44.01 &0.01 &0.00 &34.46 &0.00 &15.97 &3.59 &33.56 &0.01 &0.00 &0.00 &52.38 &15.30 &24.23 &13.82&13.17 &9.26 &6.47  &18.17\\
FCN-8s~\cite{long2015fully} &78.02&39.79&58.96 &5.32 &3.08 &49.08&12.36&26.82&15.66&49.41 &6.48 &0.00 &2.16 &62.65 &29.78 &36.63 &28.12&26.05&17.76&17.70 &28.29\\
DeepLabV2~\cite{chen2018deeplab} &84.53&56.48&65.33&29.98&19.67&62.44&30.33&51.03&40.51&69.00&22.38&11.29&20.56&70.11 &49.25 &52.88 &42.37&35.78&33.81&32.89 &41.64\\
Attention~\cite{chen2016attention} &84.00&58.87&66.78&23.32&19.48&63.20&29.63&49.70&35.23&66.04&24.73&12.84&20.41&70.58 &50.17 &54.03 &38.35&37.70&26.20&27.09 &42.92\\
Attention+SSL~\cite{gong2017look} &84.56&59.75&67.25&28.95&21.57&65.30&29.49&51.92&38.52&68.02&24.48&14.92&24.32&71.01 &52.64 &55.79 &40.23&38.80&28.08&29.03 &44.73\\
ASN~\cite{luc2016semantic} &84.01&56.92&64.34&28.07&17.78&64.90&30.85&51.90&39.75&71.78&25.57 &7.97 &17.63&70.77 &53.53 &56.70 &49.58&48.21&34.57&33.31 &45.41\\
SSL~\cite{gong2017look} &84.64&58.21&67.17&31.20&23.65&63.66&28.31&52.35&39.58&69.40&28.61&13.70&22.52&74.84 &52.83 &55.67 &48.22&47.49&31.80&29.97 &46.19\\
MMAN~\cite{luo2018macro} &84.75&57.66&65.63&30.07&20.02&64.15&28.39&51.98&41.46&71.03&23.61 &9.65 &23.20&69.54 &55.30 &58.13 &51.90&52.17&38.58&39.05 &46.81\\
SS-NAN~\cite{zhao2017self} &\textbf{88.67}&63.86 &70.12&30.63&23.92&\textbf{70.27}&33.51&\textbf{56.75}&40.18&72.19&27.68&\textbf{16.98}&\textbf{26.41}&75.33&55.24&58.93&44.01&41.87&29.15&32.64
 &47.92\\
CE2P~\cite{ruan2019devil} & 87.6 & 65.29 & 72.54 & 39.09 & 32.73   & 69.46   & 32.52 & 56.28 & 49.67 & 74.11 & 27.23   & 14.19 & 22.51 & \textbf{75.5} & \textbf{65.14} & 66.59 & \textbf{60.1}  & \textbf{58.59} & \textbf{46.63}  & \textbf{46.12}  & 53.1 \\
\midrule 
Ours & 87.4 & \textbf{67.56} & \textbf{72.63} & \textbf{42.96} & \textbf{36.65} & 69.15 & \textbf{35.35} & 55.74 & \textbf{51.13} & \textbf{74.19} & \textbf{30.49} & 15.69 & 22.61 & 75.01 & 65.04 & \textbf{67.76} & 58.51 & 57.13 & 45.79 & 45.37 & \textbf{53.8} \\
\bottomrule
\end{tabular}}
\end{table*}

\subsubsection{Comparison with the state of the art}

We compare our method with the state-of-the-art approaches on three datasets, including small-scale and large-scale ones, as presented in Table~\ref{table:comparison}. 
The score of the eyes/brows is the sum of scores of the left and right ones, and the overall score is calculated by the average score of the mouth, eyes, nose, and brows.
It is worth noting that, because the codes of some algorithms are not publicly available and the requirements of training data are different, it is hard to re-implement all the methods on every dataset. Therefore, we focus on implementing and testing the latest methods while dropping several previous methods, including Liu \et \cite{liu2017face},  Lin \et \cite{lin2019face} and Lee \et \cite{CelebAMask-HQ}. 
Specifically, the work by Lin \et \cite{lin2019face} cannot be tested on the LaPa dataset and the CelebAMask-HQ dataset, because there is a lack of fine facial segmentation bounding boxes in both datasets, which is however required by \cite{lin2019face}. Liu \et \cite{liu2017face} and Lee \et \cite{CelebAMask-HQ} are improved by follow-up works such as Wei \et \cite{wei2019accurate} and Luo \et \cite{luo2020ehanet}, thus we show the better results of Wei \et \cite{wei2019accurate} and Luo \et \cite{luo2020ehanet}.

On the small-scale Helen dataset, our method achieves comparable performance with the state-of-the-art approaches. 
On the two large-scale face parsing datasets---LaPa and CelebAMask-HQ, our method outperforms the previous methods by a large margin. 
Specifically, our model achieves improvement over Te \et \cite{te2020edge} by 1.2\% on the LaPa dataset and by 0.4\% on the CelebAMask-HQ dataset, especially on brows and eyes. 

We also visualize some parsing results of the CelebAMask-HQ dataset in comparison with competitive methods in Fig.~\ref{fig:comparison}. 
We see that our results exhibit accurate segmentation even over delicate details such as the categories of hair, earring and necklace. For example, in the first row, our model distinguishes the hand from the hair component accurately even if they are overlapped, while other methods assign the label of the hair to the hand. 
In the second row, our model generates the most consistent parsing result for the thin necklace while other results are fragmented. 
In the third row, our model separates the delicate earring from the hair, producing the finest parsing result.

\subsubsection{Performance over domain gaps}
Although there is no significant domain gap among multiple datasets, there are different aspects to discuss the domain gap \cite{luo2021category,luo2019significance}. The first is to train/test on different datasets, while the second is to explore the diversity of face poses. The evaluation results are listed in Table \ref{table:cross}.

Firstly, we conduct the cross test with respect to different datasets, including the CelebAMask-HQ and Helen datasets. 
We divide the Helen dataset into the training set (denoted as "Helen-Train") and the testing set (denoted as "Helen-Test"). 
Specifically, we utilize the model trained on the CelebAMask-HQ dataset and test directly on "Helen-Test". However, these two datasets are labeled with different categories, so we select the common facial components for testing and mark the other labels as the background. 
Consequently, finer labels such as ``Glasses” and ``Ears" are categorized as the background, which affects the performance when compared with the result of evaluating on the Helen dataset as shown in Table \ref{table:cross}. 
Besides, among the common labels, the segmentation accuracy decreases mostly on "Brows" and "Mouth".

{Secondly, we conduct the cross test between data with different face poses. 
Specifically, we detect the pitch angle of each input image, and divide the Helen dataset into two categories according to whether the pitch angle is within $0^\circ-40^\circ$. If the angle is between $0^\circ$ and $40^\circ$, we refer to the images as the "Helen-Primary" dataset, and otherwise the "Helen-Reference" dataset. 
Specifically, "Helen-Reference" contains 798 images out of 2330 images in the Helen dataset, which occupies 34\% of the complete dataset. 
As presented in Table \ref{table:cross}, the performance only drops by 1.4\% if we take "Helen-Primary" as the training dataset and "Helen-Reference" as the testing dataset. 
In comparison, when we swap the datasets, the parsing result is only 42.9\% in Overall F1 score. This is because the "Helen-Reference" dataset with large pitch angles does not provide enough information for understanding complete faces, especially for subtle mouth components.}

In addition, the cross test between data with different face poses also demonstrates the performance of AGRNet when some parts of the face missing, \eg, due to the self-occlusion, where non-front faces are self-occluded in general. The good performance of training on "Helen-Primary" and testing on "Helen-Reference" shows our method generalizes to self-occluded faces well. 
\textit{This generalization ability gives credits to the intrinsic structure embedded in the proposed graph representation.} That is, even though self-occluded data may demonstrate rather different distribution characteristics, there may be an intrinsic structure embedded in the data. Such intrinsic structure may be regarded to be better maintained from seen datasets to unseen datasets. Consequently, the graph representation learned from the face structure provides extra insights from the structure domain, in addition to the data domain, that finally enhances the generalizability of the network. 

\begin{figure}[t]
    \centering
    \includegraphics[width=0.4\textwidth]{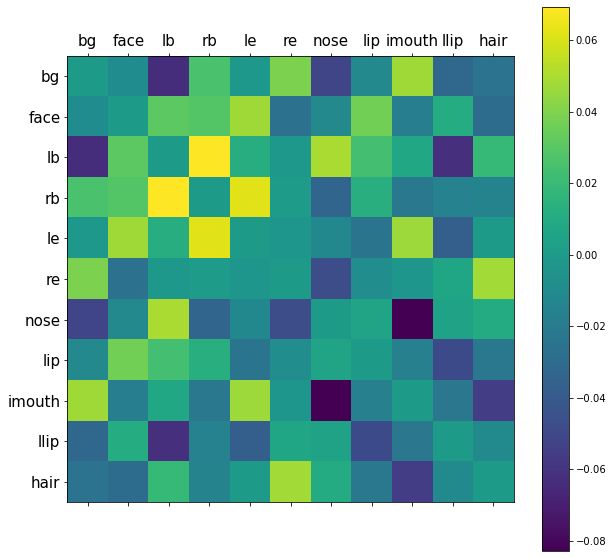}
    \vspace{-0.2cm}
    \caption{Visualization of the learned graph adjacency matrix.}
    \label{fig:graph}
\end{figure}

\subsubsection{Visualization of vertices and response}

\begin{figure}[tbp]
    \centering
    \includegraphics[width=0.43\textwidth]{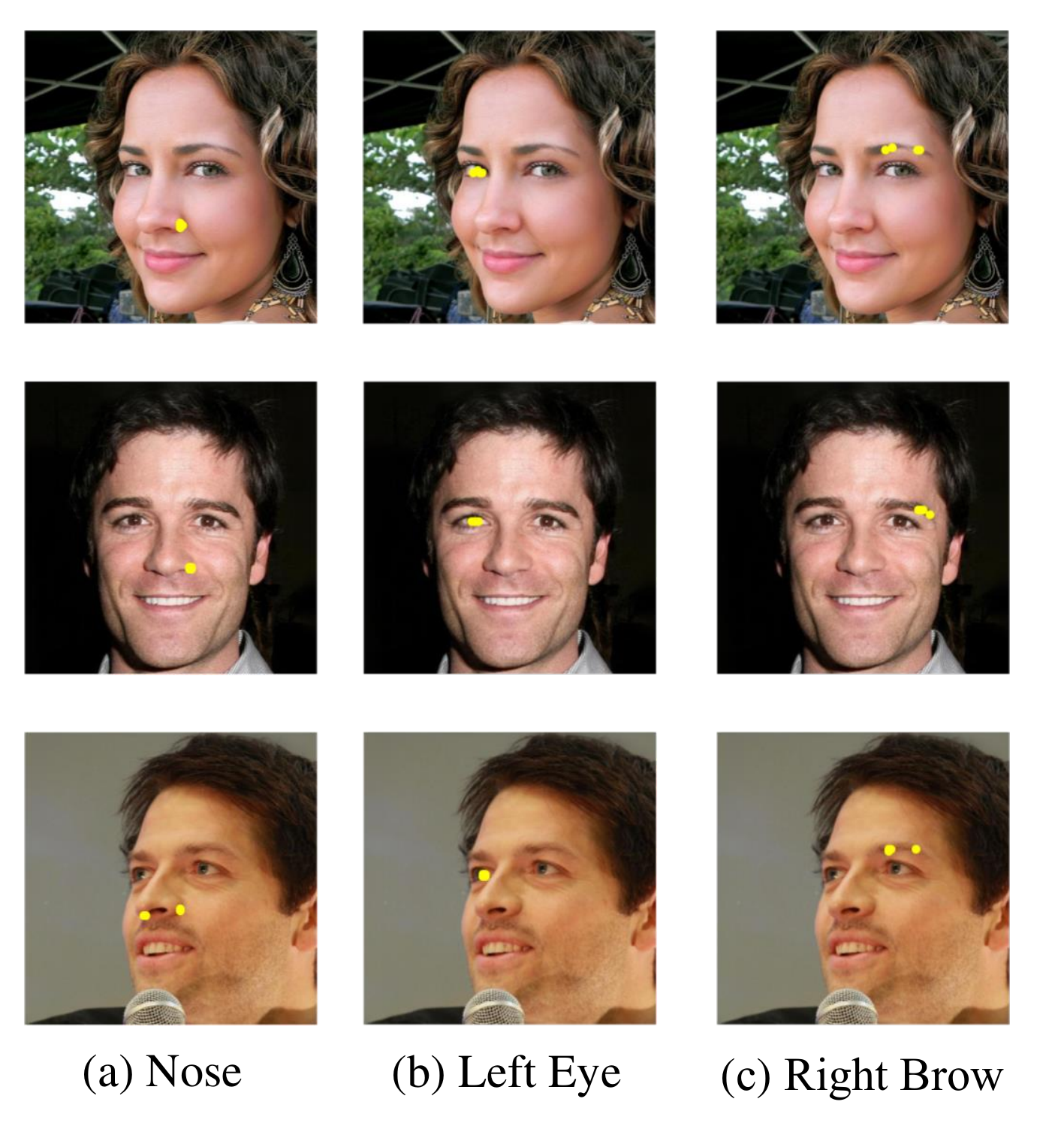}
    \caption{\textbf{Visualization of projected vertices with respect to facial components.} The yellow points represent vertices acquired from the proposed adaptive graph projection, and each column represents a certain category. Note that, the vertices significant overlap with the mainstream facial landmark layout \cite{liu2019grand, wu2018look}, which brings the advantages towards learning and reasoning the semantic representation for face parsing}
    \label{fig:anchor_vis}
\end{figure}

Further, we present some visualization examples of vertices and the pixel-to-vertex projection matrix in order to provide intuitive interpretation. Fig.~\ref{fig:anchor_vis} shows the selected vertices with respect to specific facial components, where keypoints are marked as yellow. 
We observe that vertices lie in the interior region of the corresponding facial component. 
Besides, symmetric components are well separated, such as left and right brows. 

Also, we visualize the adjacency matrix trained on the Helen dataset in Fig. \ref{fig:graph}. As shown by the color bar, lighter colors indicate larger edge weights. 
We observe that the pair of left brow ("lb" in the figure) and the right brow ("rb" in the figure) has the strongest positive correlation, while the pair of inner mouth ("imouth" in the figure) and nose has the strongest negative correlation, which is reasonable to some extent.  

\begin{figure*}[htbp]
    \centering
    \includegraphics[width=0.8\textwidth]{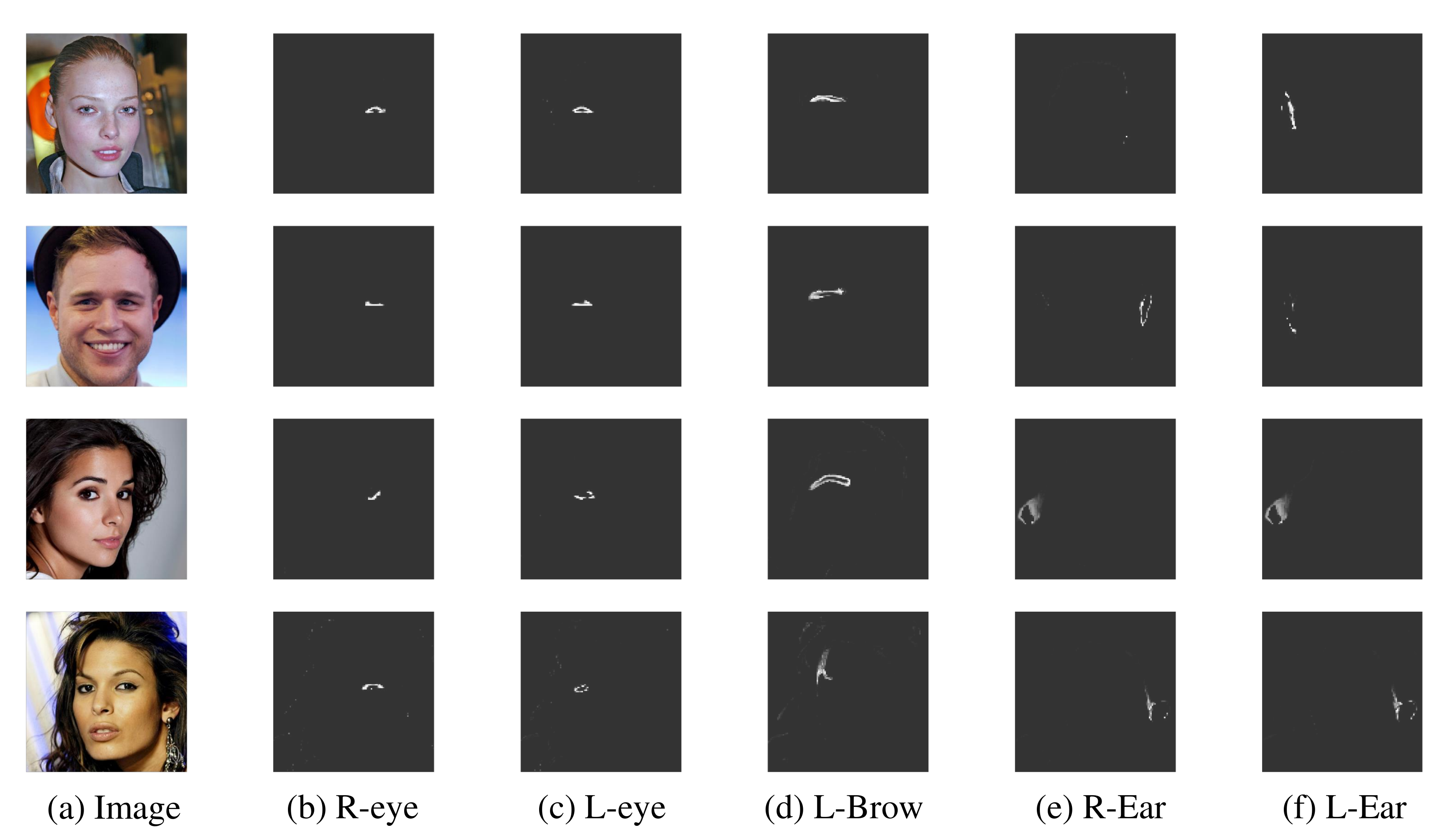}
    \caption{\textbf{Response maps with respect to different facial components.} Brighter color indicates higher response. The response maps exhibit high response to ambiguous pixels along the inter-component boundaries.}
    \label{fig:response_vis}
\end{figure*}

Furthermore, we visualize the pixel-to-vertex projection matrix via response maps. 
As in Fig.~\ref{fig:response_vis}, given a projection matrix $\P \in \mathbb{R}^{KN_c \times H_1W_1}$, we visualize the weight of each pixel that contributes to semantic vertices. Since there are 4 vertices corresponding to each component, we sum them up as a complete component response map.
Brighter color in Fig.~\ref{fig:response_vis} indicates higher response. 
We observe that pixels demonstrate high response to vertices in the corresponding face component in general, which validates the effectiveness of the proposed adaptive graph projection.   
In particular, edge pixels show higher response in each component, thanks to the proposed edge attention and Boundary-Aware loss. 
Note that, there exist outliers if some component is blocked. For example, while one ear is occluded in both images from row 3 and row 4, several pixels still exhibit high response to the other ear region in the response map.

\begin{figure}[tbp]
    \centering
    \includegraphics[width=0.35\textwidth]{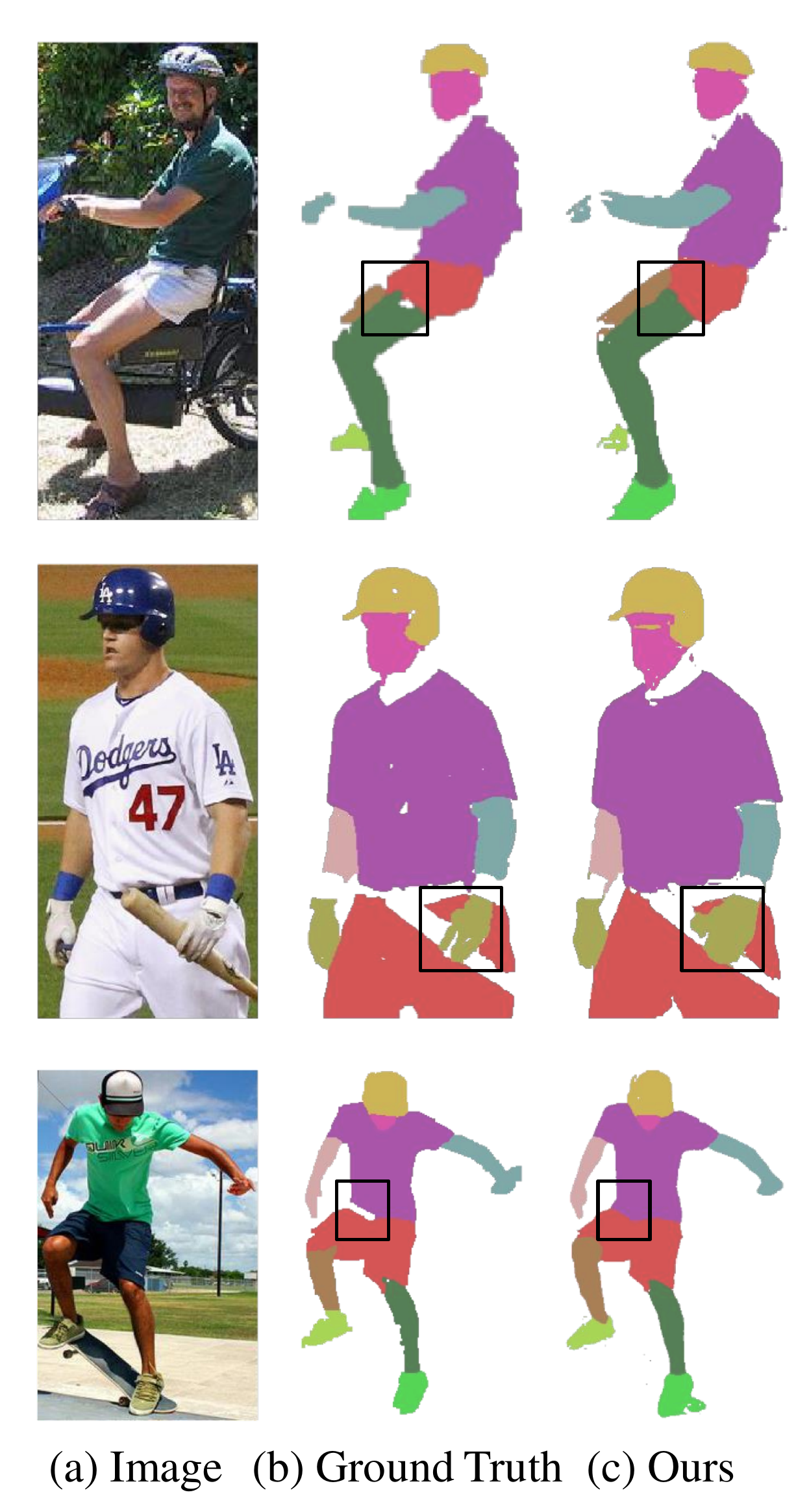}
    \caption{\textbf{Human parsing results on the LIP dataset.} Our parsing maps are clear and smooth along the boundaries, and even exhibit better results of shorts compared to the ground truth as depicted in the third row.}
    \label{fig:lip}
\end{figure}

\begin{figure}
    \centering
    \includegraphics[width=0.45\textwidth]{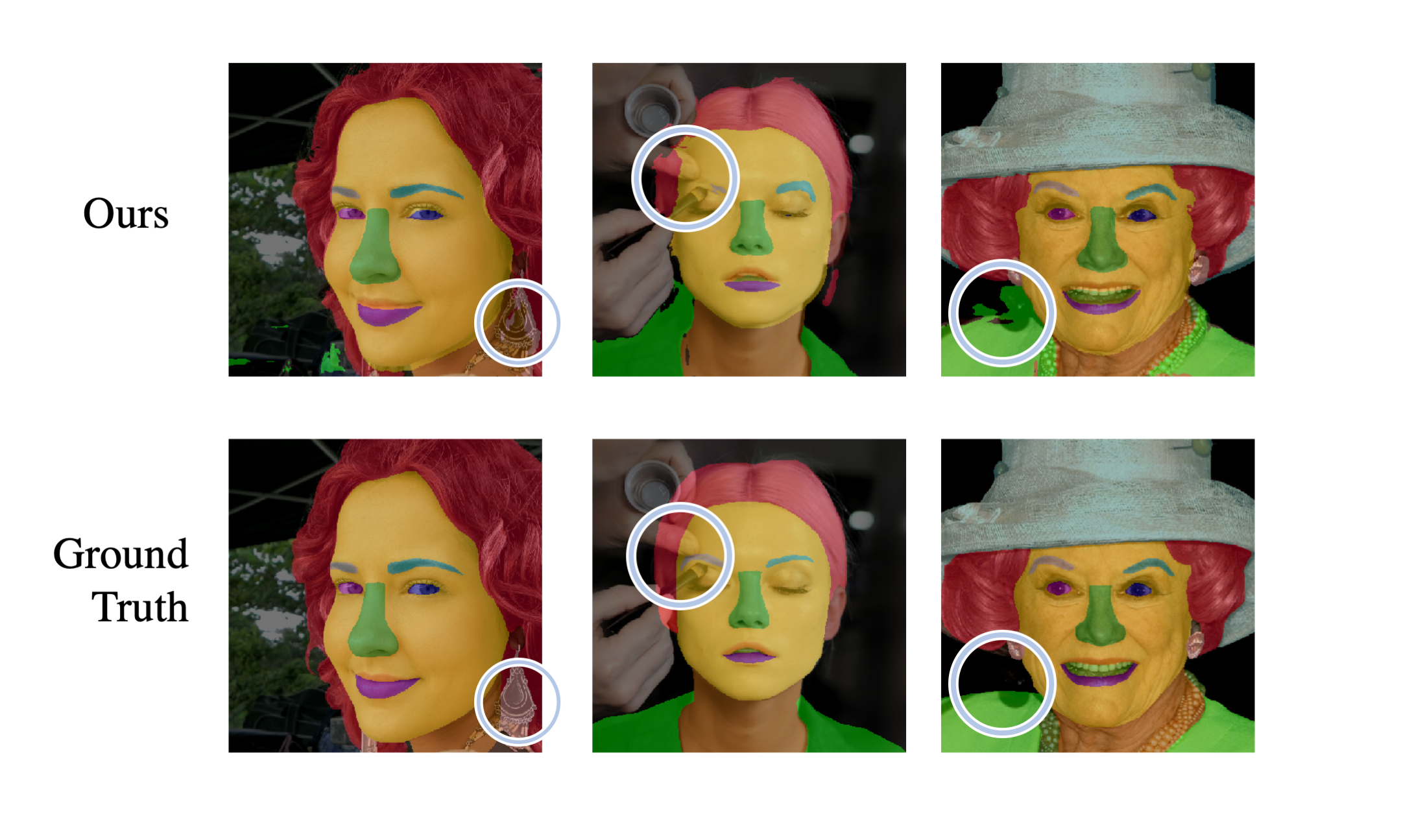}
    \caption{\textbf{Failure cases.} The earring is mislabeled in the first sample, the hair region is not continuous in the second sample, and some background is classified as nose in the third one.}
    \label{fig:failure}
\end{figure}

\subsubsection{Failure cases}
We show several unsatisfactory examples in the CelebAMask-HQ dataset in Fig.~\ref{fig:failure}, where most of incorrect labels lie along the boundaries or "accessories" categories (\eg, earring and necklace). The cause of failure cases is mostly the significant imbalance of pixel numbers in different categories or the disturbance of adjacent pixels along the boundary.

\subsection{Human Parsing}
Human parsing is another segmentation task which predicts a pixel-wise label to each semantic human part. 
Unlike face parsing, human structure is hierarchical, where a large component could be decomposed into several fragments. 
Unlike recent works which often construct a fixed graph topology based on hand-crafted human hierarchy \cite{liu2019braid,wang2019learning,wang2020learning}, our model learns the graph connectivity adaptively. 
Following the standard training and validation split as described in \cite{ruan2019devil}, we evaluate the performance of our model on the validation set of the LIP dataset, and present the results in Table~\ref{table:LIP}.
Several comparison methods involve the attention module, including SS-NAN \cite{zhao2017self} and Attention \cite{chen2016attention}. 

The results in Table~\ref{table:LIP} show that our model outperforms the competitive method CE2P \cite{ruan2019devil} by 0.7\% in mean IoU, which validates that our model is generalizable to human parsing. 
Also, it is worth noting that our model achieves significant improvement in elaborate categories, such as socks, glove and j-suits.

Furthermore, we provide visualization examples in Fig.~\ref{fig:lip}. 
The visualized results demonstrate that our model leads to accurate prediction of human components, especially around the boundaries between components, which gives credits to the proposed component-level modeling.  

%% file: 5_conclusion.tex
In this paper, we propose an adaptive graph representation learning and reasoning method (AGRNet) for the component-wise modeling of face images, aiming to exploit the component-wise relationship for accurate face parsing. 
In particular, we adaptively project the representation from the pixel space to the vertex space that represents semantic components as graph abstraction under the initial condition of a predicted parsing map. 
The image edge information is incorporated to highlight ambiguous pixels during the projection for precise segmentation along the edges.
Then, the model learns and reasons over the relationship among components by propagating features across vertices on the graph. Finally, the refined vertex features are projected back to pixel grids for the prediction of the final parsing map. 
Further, we propose a discriminative loss to learn vertices with distinct features for semantic description of facial components. Experimental results demonstrate that AGRNet sets the new state of the art on large-scale face parsing datasets, with accurate segmentation along the boundaries. 
Also, our model shows effectiveness on the human parsing task, which validates its generalizability.


%% file: bio.tex
\vspace{-0.5in}

\begin{IEEEbiography}
	[{\includegraphics[width=1in,height=1.25in,clip,keepaspectratio] {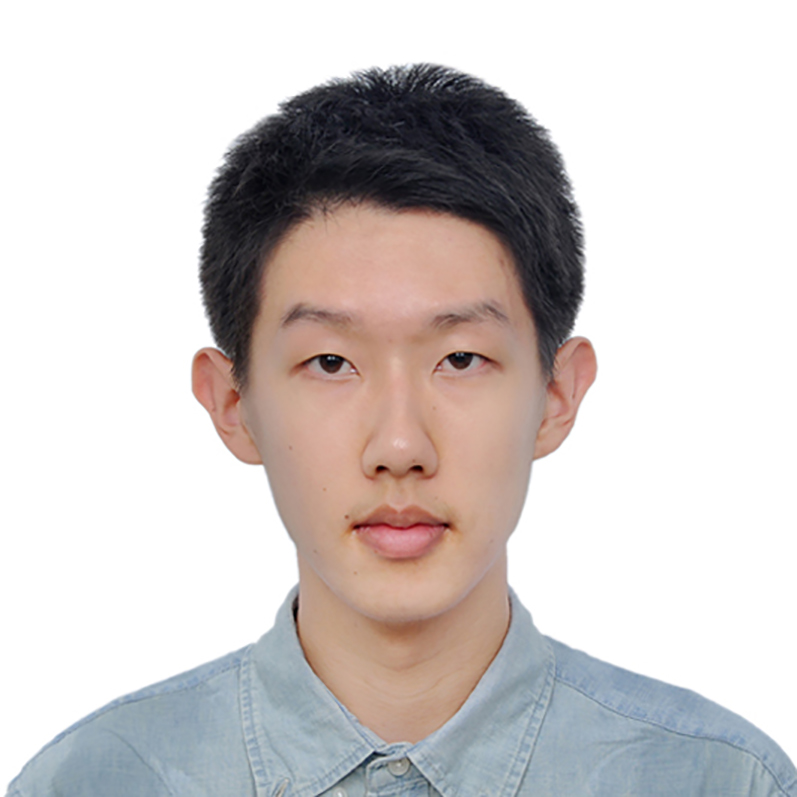}}]
	{Gusi Te}
	Gusi Te is currently an MSc student with Wangxuan Institute of Computer Technology, Peking University. Before that, he acquired his bachelor degree in Peking University. His research interests include 3D computer vision and graph neural networks.
\end{IEEEbiography}

\begin{IEEEbiography}
	[{\includegraphics[width=1in,height=1.25in,clip,keepaspectratio] {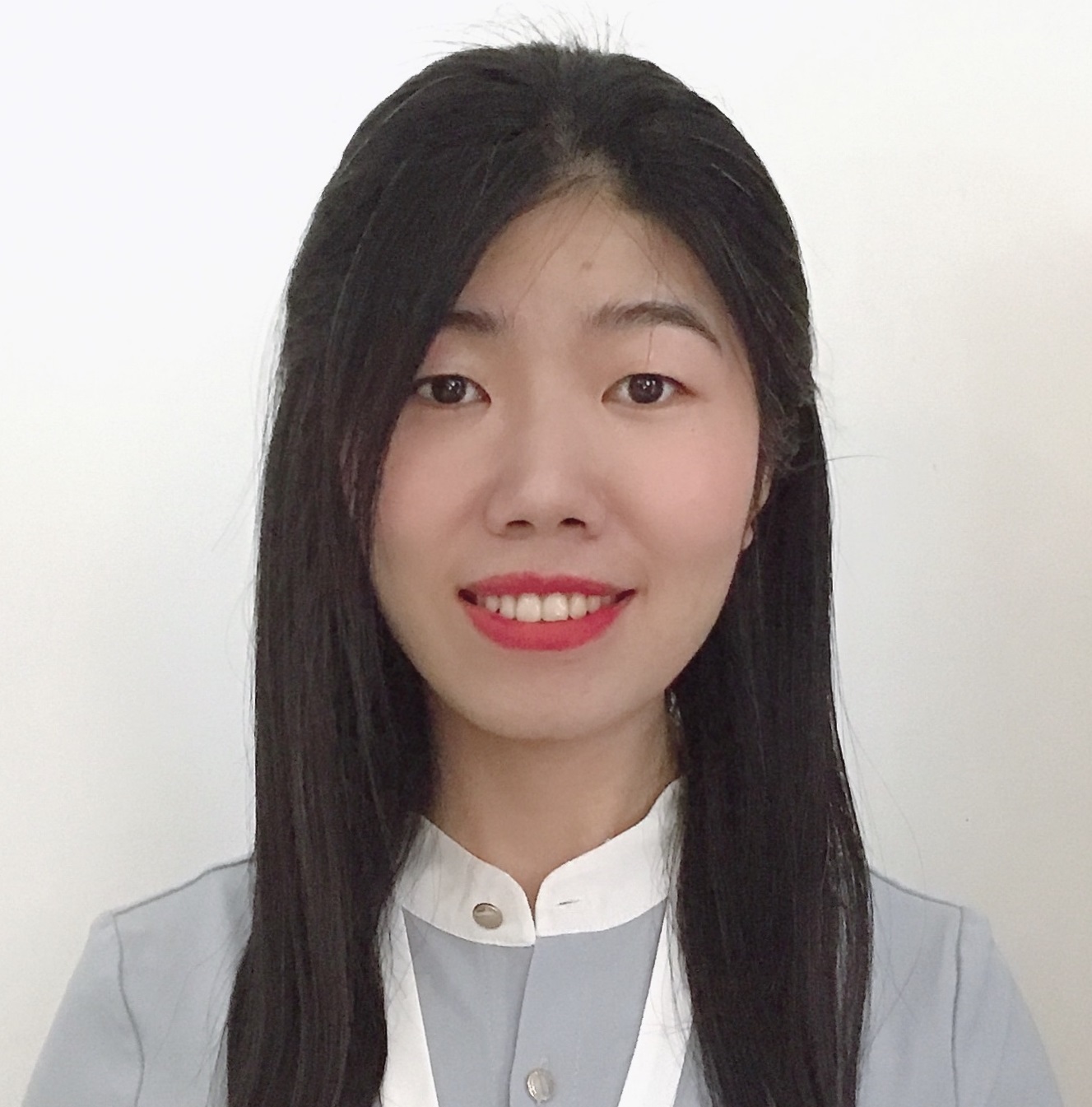}}]
	{Wei Hu}
	(Senior Member, IEEE) received the B.S. degree in Electrical Engineering from the University of Science and Technology of China in 2010, and the Ph.D. degree in Electronic and Computer Engineering from the Hong Kong University of Science and Technology in 2015.
	She was a Researcher with Technicolor, Rennes, France, from 2015 to 2017. She is currently an Assistant Professor with Wangxuan Institute of Computer Technology, Peking University. Her research interests are graph signal processing, graph-based machine learning and 3D visual computing. She has authored over 50 international journal and conference publications, with several paper awards including Best Student Paper Runner Up Award in ICME 2020 and Best Paper Candidate in CVPR 2021. 
	She was awarded the 2021 IEEE Multimedia Rising Star Award---Honorable Mention. She serves as an Associate Editor for Signal Processing Magazine, IEEE Transactions on Signal and Information Processing over Networks, etc. 
\end{IEEEbiography}

\begin{IEEEbiography}
	[{\includegraphics[width=1in,height=1.25in,clip,keepaspectratio] {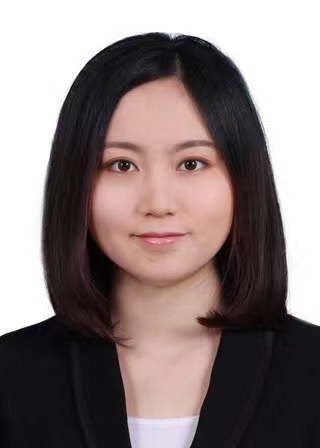}}]
	{Yinglu Liu}
    Yinglu Liu is currently a senior researcher in JD AI Research, Beijing, China. She received the B.E. degree in Information Engineering from Xiamen Unversity in 2009, and the Ph.D. degree in Pattern Recognition and Intelligent System from Institute of Automation, Chinese Academy of Sciences in 2015. Before joining JD.com, she was a senior algorithm engineer in Samsung Research China-Beijing. Her research interest includes computer vision and machine learning, with a focus on image generation, semantic segmentation, etc.
\end{IEEEbiography}

\begin{IEEEbiography}
	[{\includegraphics[width=1in,height=1.25in,clip,keepaspectratio] {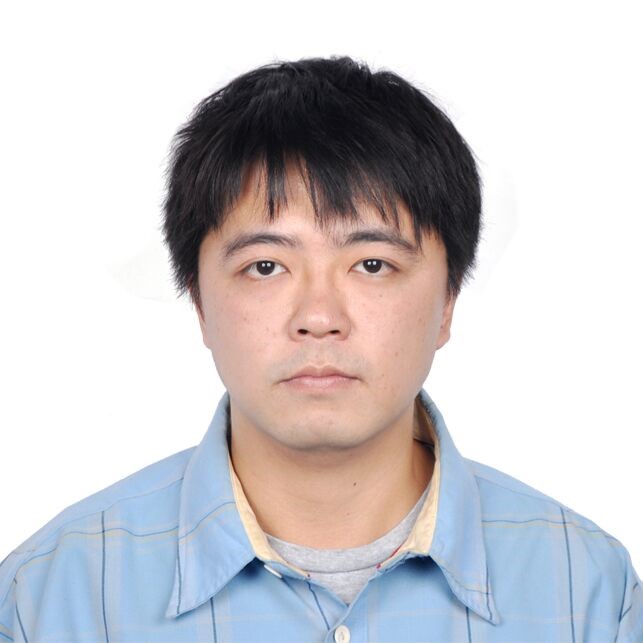}}]
	{Hailin Shi}
	Hailin Shi received his PhD from Institute of Automation, Chinese Academy of Sciences in 2017. He is currently a senior researcher at JD AI Research. His research interest includes face analysis and deep learning. He has authored or co-authored over 40 publications in the major conferences and journals of computer vision and pattern recognition, and a book chapter about deep metric learning, and has applied over 20 face-analysis-related patents. 
\end{IEEEbiography}
\begin{IEEEbiography}
	[{\includegraphics[width=1in,height=1.25in,clip,keepaspectratio] {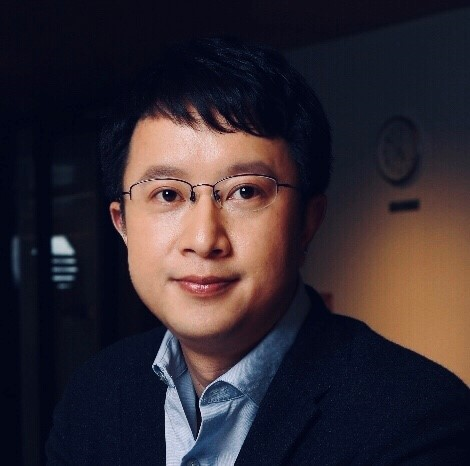}}]
	{Tao Mei}
	Tao Mei (M07-SM11-F19) is a Vice President with JD.COM and the Deputy Managing Director of JD AI Research, where he also serves as the Director of Computer Vision and Multimedia Lab. Prior to joining JD.COM in 2018, he was a Senior Research Manager with Microsoft Research Asia in Beijing, China. He has authored or co-authored over 200 publications (with 12 best paper awards) in journals and conferences, 10 book chapters, and edited five books. He holds over 25 US and international patents. He is or has been an Editorial Board Member of IEEE Trans. on Image Processing, IEEE Trans. on Circuits and Systems for Video Technology, IEEE Trans. on Multimedia, ACM Trans. on Multimedia Computing, Communications, and Applications, Pattern Recognition, etc. He is the General Co-chair of IEEE ICME 2019, the Program Co-chair of ACM Multimedia 2018, IEEE ICME 2015 and IEEE MMSP 2015. Tao received B.E. and Ph.D. degrees from the University of Science and Technology of China, Hefei, China, in 2001 and 2006, respectively. He is an adjunct professor of University of Science and Technology of China, The Chinese University of Hong Kong (Shenzhen), and Ryerson University. Tao is a Fellow of IEEE (2019), a Fellow of IAPR (2016), a Distinguished Scientist of ACM (2016), and a Distinguished Industry Speaker of IEEE Signal Processing Society (2017).

\end{IEEEbiography}